\documentclass[10pt,journal,compsoc]{IEEEtran}

\ifCLASSOPTIONcompsoc
  \usepackage[nocompress]{cite}
\else
  \usepackage{cite}
\fi

\ifCLASSINFOpdf
  \usepackage[pdftex]{graphicx}
\else
\fi

\usepackage{mathtools}
\usepackage{bm}
\usepackage{booktabs}
\usepackage{caption}
\usepackage{algorithm}
\usepackage{algorithmic}
\usepackage{multirow}

\usepackage{color}

\ifCLASSOPTIONcompsoc
  \usepackage[caption=false,font=footnotesize,labelfont=sf,textfont=sf]{subfig}
\else
  \usepackage[caption=false,font=footnotesize]{subfig}
\fi

\newcommand{\etal}{\textit{et al}.}
\newcommand{\ie}{\textit{i}.\textit{e}.}
\newcommand{\eg}{\textit{e}.\textit{g}.}

\usepackage{hyperref}
\hypersetup{
    colorlinks=true,
    linkcolor=blue,
    filecolor=magenta,      
    urlcolor=magenta,
}
\usepackage{url}

\begin{document}

\title{Learning Multi-Attention Context Graph for Group-Based Re-Identification}
%
%
%
%

\author{Yichao~Yan,
        Jie~Qin,~\IEEEmembership{Member,~IEEE},
        Bingbing~Ni,
        Jiaxin~Chen,
        Li~Liu,~\IEEEmembership{Senior Member,~IEEE},
        Fan~Zhu,
        Wei-Shi~Zheng,
        Xiaokang~Yang,~\IEEEmembership{Fellow,~IEEE},
        and~Ling~Shao,~\IEEEmembership{Senior Member,~IEEE}
\IEEEcompsocitemizethanks{\IEEEcompsocthanksitem Y. Yan, J. Qin, J. Chen, L. Liu, F. Zhu and L. Shao are with the Inception Institute of Artificial Intelligence, Abu Dhabi, United Arab Emirates. L. Shao is also with the Mohamed bin Zayed University of Artificial Intelligence, Abu Dhabi, United Arab Emirates.
E-mail: \{firstname.lastname\}@inceptioniai.org. (\textit{Corresponding author: Jie Qin})
\IEEEcompsocthanksitem B. Ni and X. Yang are with Shanghai Jiao Tong University, Shanghai, China. X. Yang is also with MoE Key Lab of Artificial Intelligence, AI Institute, Shanghai Jiao Tong University, China.
E-mail: \{nibingbing, xkyang\}@sjtu.edu.cn.
\IEEEcompsocthanksitem W.-S. Zheng is with the School of Data and Computer Science, Sun Yat-sen University, China. W.-S. Zheng is also with Peng Cheng Laboratory, Shenzhen 518005, China.
E-mail: wszheng@ieee.org.}%
}


\IEEEtitleabstractindextext{%
\begin{abstract}
Learning to re-identify or retrieve a group of people across non-overlapped camera systems has important applications in video surveillance. However, most existing methods focus on (single) person re-identification (re-id), ignoring the fact that people often walk in groups in real scenarios. In this work, we take a step further and consider employing context information for identifying groups of people, \ie, group re-id. On the one hand, group re-id is more challenging than single person re-id, since it requires both a robust modeling of local individual person appearance (with different illumination conditions, pose/viewpoint variations, and occlusions), as well as full awareness of global group structures (with group layout and group member variations). On the other hand, we believe that person re-id can be greatly enhanced by incorporating additional visual context from neighboring group members, a task which we formulate as group-aware (single) person re-id. In this paper, we propose a novel unified framework based on graph neural networks to simultaneously address the above two group-based re-id tasks, \ie, group re-id and group-aware person re-id. Specifically, we construct a context graph with group members as its nodes to exploit dependencies among different people. A multi-level attention mechanism is developed to formulate both intra-group and inter-group context, with an additional self-attention module for robust graph-level representations by attentively aggregating node-level features. The proposed model can be directly generalized to tackle group-aware person re-id using node-level representations. Meanwhile, to facilitate the deployment of deep learning models on these tasks, we build a new group re-id dataset which contains more than $3.8K$ images with $1.5K$ annotated groups, an order of magnitude larger than existing group re-id datasets. Extensive experiments on the novel dataset as well as three existing datasets clearly demonstrate the effectiveness of the proposed framework for both group-based re-id tasks. The code is available at \url{ https://github.com/daodaofr/group_reid}.


\end{abstract}

}

\maketitle

\IEEEdisplaynontitleabstractindextext

\IEEEpeerreviewmaketitle

\IEEEraisesectionheading{\section{Introduction}\label{sec:introduction}}

\begin{figure}[t]
\subfloat[Person re-id\label{subfig-1-1}]{%
   \includegraphics[width=\linewidth]{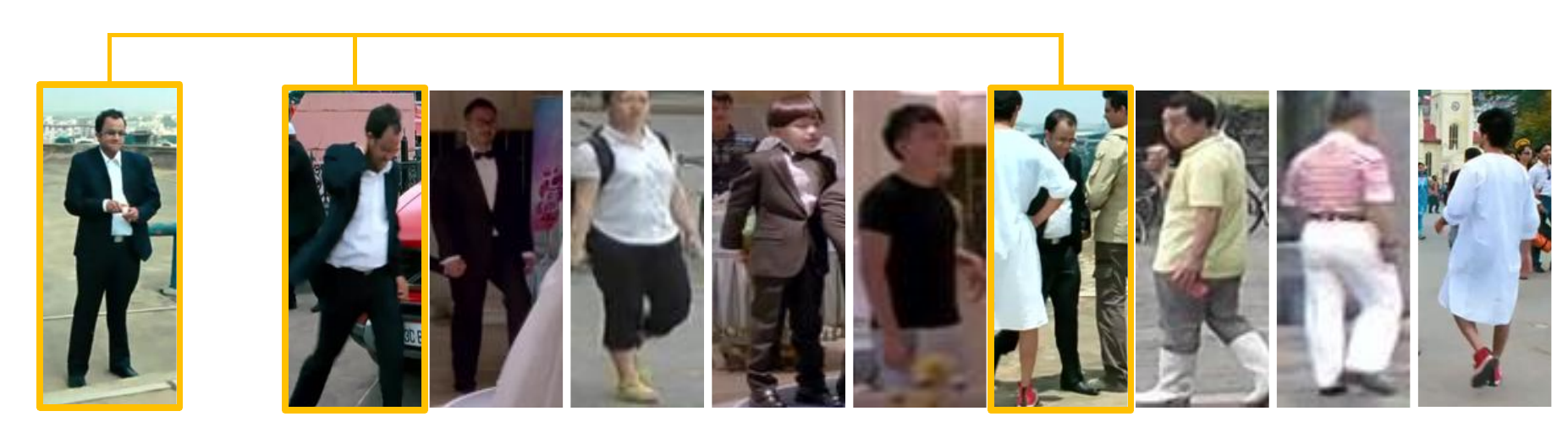}
}
 \hfill
\subfloat[Group re-id\label{subfig-1-3}]{%
   \includegraphics[width=\linewidth]{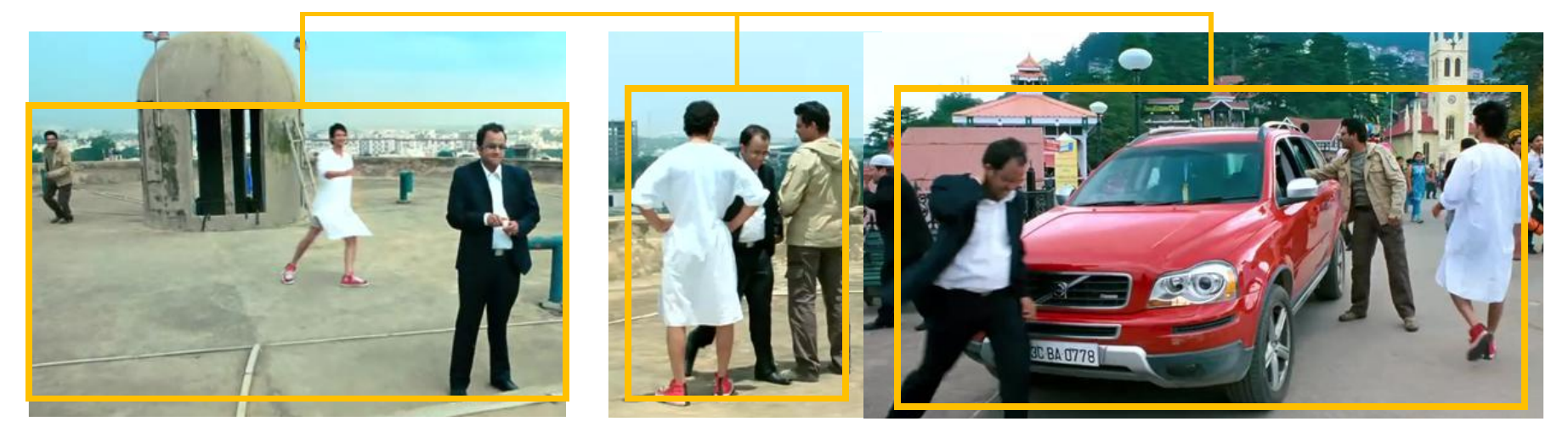}
}
 \hfill
\subfloat[Group-aware person re-id\label{subfig-1-2}]{%
   \includegraphics[width=\linewidth]{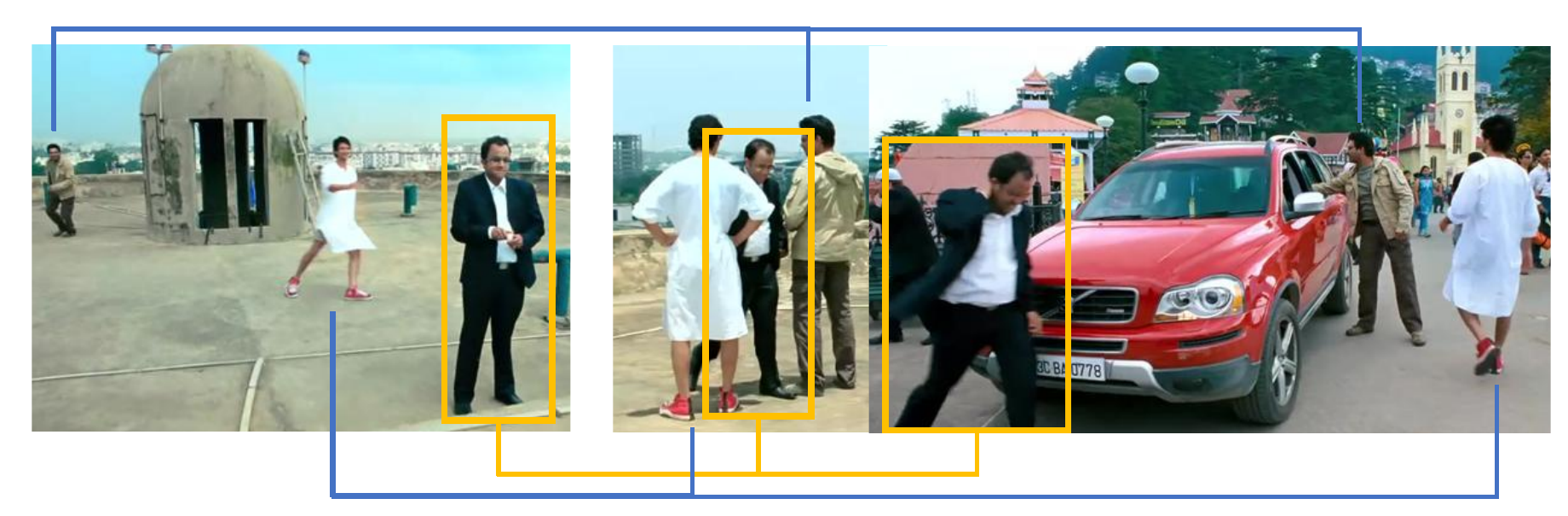}
}
 \caption{Illustration of traditional person re-id, group re-id and group-aware (single) person re-id. (a) Person re-id only measures the similarity between individual pairs, and its performance is easily influenced by occlusions or people wearing similar clothes. (b) Group re-id aims to associate different groups of people, which is more challenging compared with single person re-id. (c) In group-aware (single) person re-id, we explore the visual context information in the group as additional guidance to learn more robust representations. For instance, the man in the black suit can be better re-identified with the context information from his neighboring group members.}
 \label{fig:1}
\end{figure}

\IEEEPARstart{P}{erson} re-identification (re-id) aims to re-identify individuals across multi-camera surveillance systems. Over the past few years, person re-id has received increasing attention due to its great potential in many real-world applications, such as searching for suspects or lost people. In addition, it is also a fundamental research topic in computer vision and pattern recognition. In a typical person re-id pipeline, the system is provided with a target person as probe and aims to search through a gallery of known ID records to find a match. Usually, the probe and the gallery consist of human detection results or manually annotated bounding boxes, as shown in Fig.~\ref{subfig-1-1}.
The major challenges in person re-id include distinguishing different people sharing a similar appearance (\eg, wearing similar clothes), or, conversely, retrieving the same person that undergoes significant appearance changes (due to pose/viewpoint variations, different illumination conditions, and occlusions). Although sophisticated deep learning architectures and metric learning schemes have greatly improved the individual representation capability, the conventional setting for person re-id neglects to take into account the fact that people are likely to walk in groups in real scenarios.

Based on the above fact, in this paper, we aim at solving two group-based re-identification problems, \ie, group re-id and group-aware (single) person re-id. As shown in Fig.~\ref{subfig-1-3} and \ref{subfig-1-2}, respectively, group re-id aims to find target groups given a group of interest, whilst group-aware person re-id is essentially a person re-id task, differing only in the fact that it considers context information from neighboring group members. We believe that, on the one hand, group re-id can greatly facilitate the understanding of group behaviors; while, on the other hand, rich context within a group can also be beneficial for single person re-id. However, there are very limited existing works~\cite{DBLP:conf/bmvc/ZhengGX09,DBLP:conf/icpr/CaiTP10,DBLP:conf/eccv/RistaniSZCT16,DBLP:conf/iccv/LisantiMBF17,xiao2018group} that address group-based re-id tasks, which is mainly due to the following challenges. \emph{First of all}, in addition to the difficulties already faced in person re-id, multiple subjects in a group bring more challenges, which conventional person re-id models cannot directly deal with. A straightforward solution may be to pool or average individual features, but this is not a robust solution since it does not allow groups with similar centroids to be distinguished. \emph{Second}, group members are not always spatially close to each other (see Fig.~\ref{subfig-1-3}), and the group layout can vary greatly depending on the viewpoint. Therefore, directly utilizing the contents in group bounding boxes for re-id is not feasible. \emph{Third}, the number of group members varies for different groups, and even for the same group may change across frames. Learning a robust group-level representation with such data discrepancies is a non-trivial problem. \emph{Last but not least}, most previous methods employ hand-crafted features for group representation, which goes against current trends of employing deep learning models. This is probably due to the lack of large-scale group re-id datasets. 
Therefore, it is highly desirable to develop a deep learning based model that not only addresses the above challenges for group re-id, but also makes good use of the context information in groups to boost the performance of person re-id.

\begin{figure*}
    \centering
    \includegraphics[width=\linewidth]{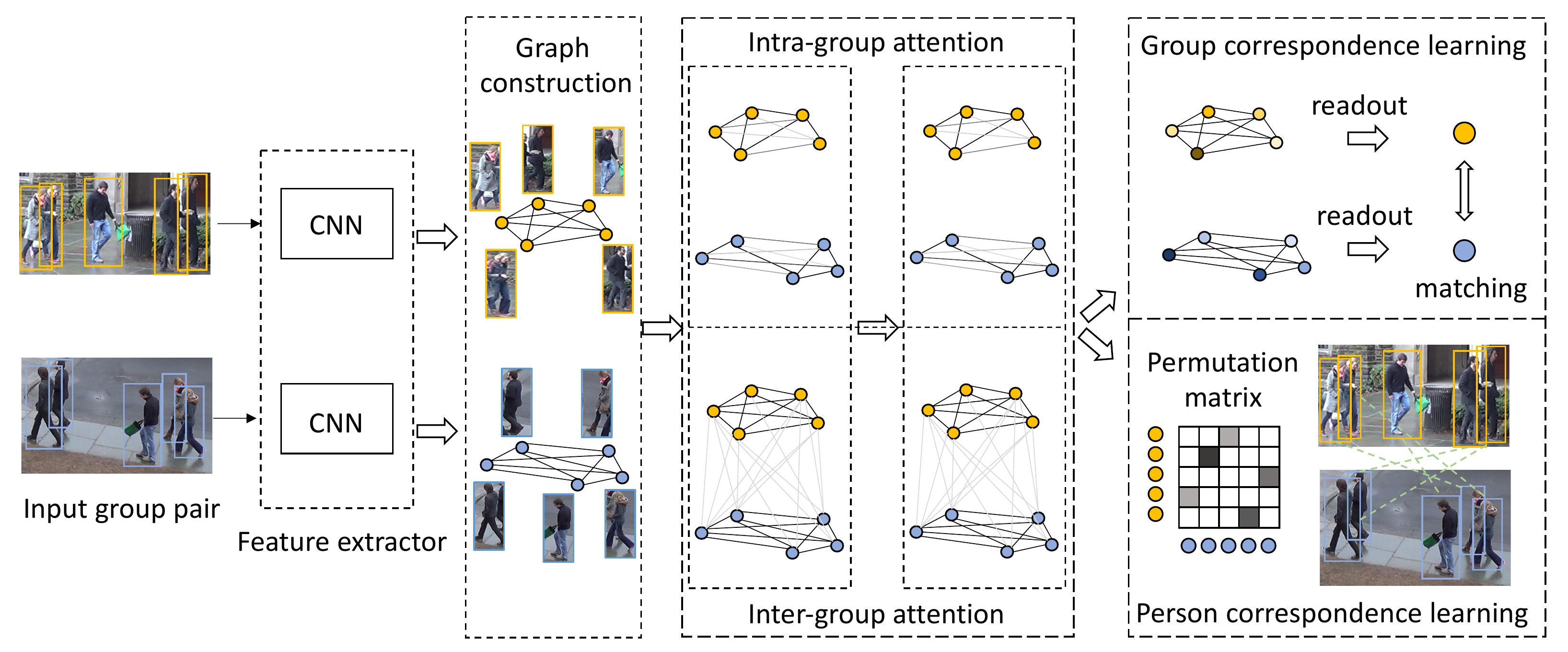}
    \caption{Illustration of the proposed Multi-Attention Context Graph for group-based re-identification. First of all, the individual features are extracted by a CNN model for group member representation. Second, we construct a context graph for each group by connecting all the group members with graph nodes. We then design an intra-graph and an inter-graph attention module to learn the within-group and between-group dependencies, respectively. Afterwards, node-level features are aggregated for global group-level and local person-level correspondence learning, respectively. }
    \label{fig:pipeline}
\end{figure*}

To address the aforementioned challenges, we propose a novel framework named \textbf{Multi-Attention Context Graph (MACG)}, which is a unified model for both group re-id and group-aware person re-id. Specifically, as shown in Fig.~\ref{fig:pipeline}, we model each group as a context graph, where the nodes refer to group members. The advantages of our context graph representation are two-fold. First, a graph-level representation can be obtained through feature aggregation from nodes, which inherently addresses the challenge of group layout and/or group member changes. Second, node features in the group can benefit from the context information through the use of Graph Neural Networks (GNNs), which greatly facilitate information propagation through graph edges. In this way, group re-id can be formulated as a graph-level feature learning task. Furthermore, we propose a multi-level attention mechanism to address the challenges of learning group context information. For the node-level representation, intra- and inter-graph attention modules are proposed for encoding the context information within the same graph and across different graphs, respectively. These modules are partially inspired by the recent attention mechanisms for GNNs (\eg, GAT~\cite{DBLP:journals/corr/abs-1710-10903} and GAM~\cite{DBLP:conf/kdd/LeeRK18}). We further build a higher-level attention mechanism for aggregating node-level features to obtain the final graph-level representation. This module adaptively finds the most representative individuals within the group and thus increases the discrimination capability of group-level features. Note that all three attention modules work collaboratively and are optimized jointly. In practice, graph-level representations are directly employed for group re-id, while node-level representations can be leveraged to address group-aware person re-id. 

In addition, as previously mentioned, the lack of large-scale datasets for group re-id significantly hinders algorithmic development (and especially the development of deep learning approaches) in this important research field. As a matter of fact, the current largest group re-id dataset~\cite{xiao2018group} only contains 177 groups from 354 images, causing the learned models to easily over-fit to the training data. To facilitate the deployment of deep learning models on this task and to improve the generalizability of the learned models, we build a new group re-id dataset containing \textbf{3,839} group images of \textbf{1,558} annotated groups, which is an order of magnitude larger than existing datasets. Since this dataset is built upon CUHK-SYSU~\cite{DBLP:conf/cvpr/XiaoLWLW17}, which was originally built for the person search task, we denote the novel dataset as \textbf{CUHK-SYSU-Group (CSG)}.

The main contributions of this work include
\begin{enumerate}
\item We build a novel dataset, \ie, CUHK-SYSU-Group (CSG), which is currently the largest group re-id dataset. We believe the dataset will significantly boost the research in this important field.
\item We propose a novel unified framework, namely MACG, for both group re-id and group-aware person re-id tasks. Context information exploited by GNNs largely enhances person(node)-level and group(graph)-level representation capabilities.
\item A multi-level attention mechanism is developed to capture both intra- and inter-group context. Moreover, the final graph-level representation is learned from node-level features in a self-attentive way.
\item The proposed framework is systematically evaluated on the novel CSG dataset as well as three existing group re-id datasets, where the experimental results clearly show that our MACG outperforms the state-of-the-art methods by a large margin in terms of both group-based re-id tasks.
\end{enumerate}

The remainder of this paper is organized as follows. Section~\ref{sec:related} introduces some related works. The proposed MACG framework is present in Section~\ref{sec:method}. Section~\ref{sec:data} elaborates the details of the newly collected CSG dataset, including the annotation rule and the statistics. In Section~\ref{sec:experiments}, the performance of MACG on CSG as well as other three exiting group re-id datasets, is evaluated by comparing to the state-of-the-art methods. Finally, we draw some conclusions and summarize the future research direction in Section~\ref{sec:conclusion}.


\section{Related Work}\label{sec:related}
\textbf{Person Re-Identification}. Person re-id aims to associate pedestrians across non-overlapping cameras. Most previous methods try to address this task from two perspectives, \ie, feature representation and distance metric learning. 
Before the emergence of deep learning methods, previous approaches resorted to designing various types of hand-crafted features, such as color~\cite{DBLP:journals/ijcv/Lowe04}, texture~\cite{DBLP:conf/cvpr/FarenzenaBPMC10}, and gradient~\cite{DBLP:conf/eccv/GrayT08,BEDAGKARGALA20121908}. These methods achieved certain success on small datasets. However, the representation capability of hand-crafted features is limited for large-scale searches.
Similar limitations also apply for traditional distance metric learning methods~\cite{6226421,DBLP:conf/cvpr/ZhangXG16,DBLP:conf/cvpr/KostingerHWRB12,CHENG2018}, which aim to optimize a distance function based on certain features. However, the learned metric usually over-fits on the training data, thus leads to limited generalization ability. 

With its recent renaissance~\cite{DBLP:conf/eccv/LuMNYRY18,DBLP:journals/tip/WangSS18}, deep learning was first introduced to address person re-id in~\cite{DBLP:conf/cvpr/ZhaoOW14,DBLP:conf/cvpr/LiZXW14}, and has largely dominated this field ever since. A large number of works have proposed different model structures for this task~\cite{DBLP:conf/cvpr/ZhaoOW14,DBLP:conf/cvpr/LiaoHZL15,DBLP:conf/eccv/YanNSMYY16,DBLP:conf/cvpr/ChenYCZ16,DBLP:conf/mm/SongNYRXY17,DBLP:conf/cvpr/ZhouHWWT17,DBLP:journals/pami/ZhengGX16,DBLP:conf/cvpr/BaiBT17,DBLP:journals/corr/abs-1901-10650}. Some methods employ end-to-end feature and metric learning~\cite{DBLP:conf/cvpr/AhmedJM15,DBLP:conf/cvpr/SchroffKP15,DBLP:conf/cvpr/ChenCZH17,ZHAO2018,DBLP:journals/corr/ZhengZY16,7849147,7393860,chen2015relevance,chen2017fast}, while others exploit part information to build more robust representations~\cite{DBLP:conf/iccv/ShenLYXWW15,DBLP:journals/tip/LinSYXWWL17,DBLP:conf/iccv/SuLZX0T17,DBLP:conf/mm/WeiZY0T17,DBLP:journals/corr/ZhengHLY17,DBLP:conf/iccv/MiaoWLD019,chen2020acmmm}. Recently, generative adversarial networks (GANs)~\cite{DBLP:conf/nips/GoodfellowPMXWOCB14} have begun to be employed in re-id models~\cite{Zheng_2017_ICCV,DBLP:conf/cvpr/LiuNYZCH18,DBLP:conf/cvpr/WeiZ0018,DBLP:conf/eccv/QianFXWQWJX18,DBLP:conf/nips/GeLZYYWL18} to address the lack of training data and domain adaptation for this task. 
Based on the learned features, various re-ranking strategies~\cite{DBLP:conf/cvpr/ZhongZCL17,Bai_2019_CVPR} have further been proposed to refine the retrieval results. Recently, person search methods~\cite{DBLP:conf/cvpr/XiaoLWLW17,DBLP:conf/iccv/LiuFJKZQJY17,DBLP:conf/eccv/ChenZOYT18,DBLP:conf/eccv/ChangHSLYH18,DBLP:conf/cvpr/MunjalATG19} have been proposed to jointly address the task of person detection and re-identification, facilitating real-world applications. 
These approaches have achieved promising results on recent person re-id benchmarks. 
However, they all focus only on learning individual appearance features based on given human bounding boxes. When the person appears in a group, visual context information provided by this group is ignored. In this work, we address two group-based re-id tasks by exploiting the group context information. 

\textbf{Group Re-Identification}. Existing group re-id datasets~\cite{DBLP:conf/bmvc/ZhengGX09,xiao2018group,DBLP:conf/eccv/ChoiCPS14} are essentially small-scale, which greatly hinders the research on this task. As a result, there are currently only a few works~\cite{DBLP:conf/bmvc/ZhengGX09,DBLP:conf/icpr/CaiTP10,DBLP:conf/eccv/RistaniSZCT16,DBLP:conf/iccv/LisantiMBF17,xiao2018group} addressing the group re-id task, most of which are based on hand-crafted features. In this work, we build a large-scale group re-id dataset to facilitate group feature learning. Employing group information could be a promising direction to further improve single person re-id. Although some methods~\cite{DBLP:conf/eccv/AssariIS16,DBLP:conf/iccvw/CaoCHP17} have made efforts towards this for group-aware person re-id, we build a novel graph-based context learning framework upon GNNs, which effectively makes use of group context cues to improve the model.

\textbf{Graph Neural Networks}. 
Graph Neural Networks (GNNs)~\cite{DBLP:journals/tnn/ScarselliGTHM09} allow graph representations to be learnt with neural networks.
GNNs were originally developed for structured data~\cite{DBLP:conf/nips/DuvenaudMABHAA15,DBLP:conf/icml/NiepertAK16,DBLP:conf/nips/DefferrardBV16,DBLP:conf/miccai/KtenaPFRLGR17,DBLP:conf/eccv/QiWJSZ18}, but have recently been generalized to various computer vision tasks. For instance, Wang \etal ~\cite{DBLP:conf/eccv/WangG18} built a similarity graph and a spatial-temporal graph to model intra-object and inter-object relations, respectively, in a video, which achieves great performance in action recognition.
Chen \etal~\cite{Chen_2019_CVPR} applied GNNs to multi-label image recognition. They built a graph over object labels and utilized GNNs to map the label graph to a set of object classifiers, obtaining image-level labels.
Zheng \etal~\cite{Zheng_2019_CVPR} formulated the visual dialog task as the inference of the unknown nodes in a graphic model. They modeled the given question as known nodes in a graph, while the answer was formulated as a node with a missing value, which can be approximated with differentiable GNNs. Some recent works~\cite{DBLP:conf/iccv/WuLYLLL19,DBLP:conf/eccv/ShenLYCW18,DBLP:conf/cvpr/ShenLXYCW18} explored GNNs for the task of person re-id, without considering group information. 
Although graph-based models typically employ the information from neighbor points, the critical idea is to select suitable neighbors and to pass specific messages in a proper way according to the characteristics of the specific task. In this sense, our framework is clearly different from existing graph methods. For example, shape index~\cite{DBLP:conf/cvpr/RenCW014} addresses the task of facial landmark detection, and the locations of individual landmarks depend on the local pixel context. In this case, the neighbors are defined as the points within a local region around the predicted landmark, and the messages are passed by calculating pixel difference features. In contrast, our task is to measure the similarity between groups, which is dependent on the similarity between individual persons between groups. Therefore, in this work, we define the neighbors according to inter-group similarity, and we build multi-attention GNNs to model the rich context and to better propagate the node features. 

\textbf{Attention Mechanisms}.
The attention mechanism~\cite{DBLP:journals/corr/BahdanauCB14} was originally introduced in machine translation to automatically search for the most relevant parts in the source sentence and predict a target word. In computer vision tasks, attention mechanisms were designed to discover the important spatial regions in an image or the critical frames in a video. They have been broadly applied in various tasks, such as image recognition
~\cite{DBLP:journals/corr/BaMK14,DBLP:conf/mm/YanNY17,DBLP:conf/cvpr/WangJQYLZWT17}, captioning~\cite{DBLP:conf/icml/XuBKCCSZB15,DBLP:conf/cvpr/LuXPS17,DBLP:conf/iccv/YaoTCBPLC15}, and person re-id~\cite{DBLP:conf/cvpr/LiZG18,DBLP:conf/cvpr/Xu00WO18,fu2019sta}. They have also been widely employed in graph models to adaptively learn the node weights for feature propagation~\cite{DBLP:journals/corr/abs-1710-10903,DBLP:journals/corr/abs-1710-09599,DBLP:journals/corr/abs-1803-03735,DBLP:conf/aaai/Han0S18,BAI2021107637}. In this work, we build both intra-graph and inter-graph attention modules to learn the contribution of context nodes, as well as a readout attention module for node feature aggregation. These three attention modules are trained in a collaborative way for robust feature learning.

Note that the preliminary version of this work~\cite{Yan_2019_CVPR} was published at CVPR 2019. Compared with \cite{Yan_2019_CVPR}, in this paper we provide the following additional contributions. 1) The preliminary work only considers group-aware person re-id, while this work further addresses group re-id in a unified framework. 2) The preliminary work only employs a vanilla GNN model for group context mining, while this work incorporates a multi-level attention mechanism for more discriminative feature learning. 3) Experiments in~\cite{Yan_2019_CVPR} are based on CUHK-SYSU, which provides no explicit group information.
In this paper, we have further manually annotated and purified the group information in this dataset. Consequently, a new large dataset, containing much more labeled group information than existing benchmarks, is constructed, which can benefit deep  models in learning group context information more accurately and robustly.

\section{Multi-Attention Context Graph}\label{sec:method}
\subsection{Overview of the Method}
Group re-identification is a very challenging and complicated task, since it not only suffers from difficulties occurring in single person re-id (\ie, occlusions, human pose variations, background clutter, etc.), but is also confronted with new problems such as group layout and group member variations. Therefore, the keys to re-identifying a group are two-fold. First, the model should be able to understand individual group members, \ie, to address the single person re-id task. Second, the model should also be able to build a layout-invariant and noise-insensitive group-level representation for robust group re-id.

To address these challenges, we propose a novel framework, namely Multi-Attention Context Graph (MACG), for addressing group-based re-id tasks. Overall, MACG can be regarded as a unified framework that is capable of 1) extracting discriminative features of individual group members, 2) updating individual features using context information within and between groups, and 3) achieving group-level representations for group re-identification. The pipeline of our framework is illustrated in Fig.~\ref{fig:pipeline}. Specifically, the model receives pairs of groups as input, and then extracts individual features from the members of each group. Subsequently, we construct a fully connected context graph to model the characteristics of each group. Node features of the context graph are updated with intra-graph and inter-graph information through two attention mechanisms, and group-level information is obtained by aggregating node features with another self-attention mechanism. The detailed architecture is elaborated in the following subsections.

\subsection{Individual Representation}\label{sec:id_rep}
\begin{figure*}[t]
    \centering
    \includegraphics[width=\linewidth]{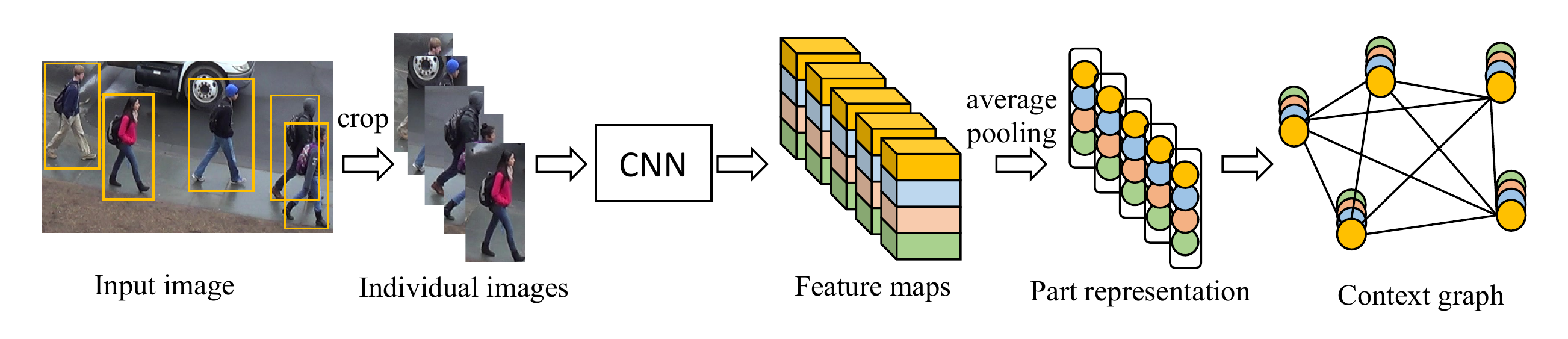}
    \caption{Illustration of individual feature extraction and context graph construction. We employ a part-based representation for each person, and part-level features are stacked as input node features in the graph. All the graph nodes are connected through edges, which makes the graph insensible to the group layout and spatial dependency within the group.}
    \label{fig:build_graph}
\end{figure*}

Individual person representation has been extensively studied in the context of single person re-id. Among the state-of-the-art methods, part-based models have been proven to be effective for robust person representation, especially in the case of partial occlusion. In our problem, people within a group are more likely to be occluded by other group members, which motivates us to also consider modeling human parts for our group re-id task. To this end, we design a part-based learning framework to effectively model part features. Specifically, we evenly divide the human body into $P$ parts ($P$=4 in our experiments). We construct several part-based pooling layers on top of the feature map from the last convolutional layer (conv5\_3) of ResNet-50~\cite{DBLP:conf/cvpr/HeZRS16}. Each part-based pooling layer concentrates on a specific human part and pools the features (\ie, a specific region in the CNN feature maps) into 2048-dimensional vectors.
These features are further utilized to construct our context graph for the subsequent graph-based feature learning procedure. The pipeline for individual feature extraction is illustrated in Fig.~\ref{fig:build_graph}.

\subsection{Context Graph Construction}
Person-level and part-level features of each group member only provide local individual representations within a group. For group re-id involving multiple people, it is more important to capture global context by aggregating local information into a global group-level representation. A trivial solution to this problem could be to use an ad-hoc feature aggregation method (such as average/max pooling). However, this may lead to significant loss of discriminative information. As an alternative, one could also employ learning-based aggregation methods, such as Recurrent Neural Networks (RNNs), but these models are unable to capture the structural dependencies in the scene. Therefore, the discrimination capability of the group representations generated by such aggregation methods is limited.

In this work, inspired by the recent success of GNNs, we propose a novel context graph model for group re-id. In GNNs, a graph is constructed to connect individual features and model the dependencies within the graph. These models can be viewed as flexible tools for both global representation learning (\eg, graph classification) and local representation learning (\eg, node classification). Specifically, in our context graph model, we first associate each group member with a graph node; thus the node-level features are the individual feature representations of the corresponding members. Second, we build a fully connected graph based on all the members in a group, in which all the nodes are connected to each other with equally weighted edges. Note that the graph could also be built to reflect the group layout, \ie, by only connecting the members that are spatially close to each other (or connecting them with edges of larger weights). However, our aim is to learn the identity information of the group, which should be invariant to the group layout. Therefore, edges with identical weights are more suitable for our task. Based on the context graph, we can perform reasoning  in a straightforward manner, with a designated graph architecture, for which this work takes advantage of graph neural networks with attention mechanisms. The learned node-level representation can be naturally aggregated into a group-level representation for group re-id. Meanwhile, the learned node-level features are also beneficial for single person re-id.

\begin{table}[t]
    \centering
        \caption{Descriptions on key notations used in this paper.}
    \label{tab:notation}
    \begin{tabular}{c|l}
    \toprule[1pt]
        $I_s$ & $s$-th input group image \\
        $I_{si}$ & Cropped image of the $i$-th person in $I_s$ \\
        $N_s$ & Number of people in $I_s$ \\
        $P$ & Number of person parts \\
        $\mathcal{G}_s$ & Graph of group context in $I_s$ \\
        $\mathbf{A}_s$ & Adjacency matrix of $\mathcal{G}_s$ \\
        $\mathcal{V}_s$ & Vertex set of $\mathcal{G}_s$ \\
        $\mathcal{E}_s$ & Edge set of $\mathcal{G}_s$ \\
        \hline
        $\mathbf{h}_{si}^{(t)}$ & Feature vector of person $i$ in the $t$-th layer of $\mathcal{G}_s$ \\
        $\mathbf{h}_{sip}^{(t)}$ & $p$-th part feature vector of person $i$ in the $t$-th layer of $\mathcal{G}_s$ \\
        $\mathbf{m}_{sip}^{(t)}$ & $p$-th intra-part message of person $i$ in the $t$-th layer of $\mathcal{G}_s$ \\
        $\mathbf{n}_{sip}^{(t)}$ & $p$-th inter-part message of person $i$ in the $t$-th layer of $\mathcal{G}_s$ \\
        $\bm{\mu}_{sip}^{(t)}$ & $p$-th inter-graph message of person $i$ in the $t$-th layer of $\mathcal{G}_s$ \\
        $\mathbf{M}$ & Affinity matrix between graphs \\
        $\mathbf{S}$ & Permutation matrix denoting node correspondence \\
    \bottomrule[1pt]
    \end{tabular}
\end{table}

Formally, given a group image $I_s$ containing $N_s$ members $\{I_{s1},...,I_{sN_s}\}$, we extract individual features $\mathbf{h}_{si}^{(0)}$ of each group member with the CNN model proposed in Sec.~\ref{sec:id_rep}:
\begin{equation}\label{eq:1}
    \mathbf{h}_{si}^{(0)} = {\rm CNN}(I_{si}), i=1,...,N_s,
\end{equation}
where $\mathbf{h}_{si}^{(0)}=[\mathbf{h}_{si1}^{(0)},...,\mathbf{h}_{siP}^{(0)}]$, and $\mathbf{h}_{sip}^{(0)}$ ($p=1,...,P$) denotes the feature of the $p$-th part of the $i$-th individual in the image $I_{s}$. The individual features serve as the input features for the graph nodes. We then build a graph $\mathcal{G}_s=\{\mathcal{V}_s,\mathcal{E}_s\}$ consisting of $N_s$ vertices $\mathcal{V}_s$ and a set of edges $\mathcal{E}_s$. 
We use $\mathbf{A}_s \in \mathcal{R}^{N_s\times N_s}$ to denote the adjacency matrix associated with graph $\mathcal{G}_s$, where we set
\begin{equation}\label{eq:2}
    \mathbf{A}_{s ({i,j})} = 1, \forall \ i,j \in \{1,...,N_s\}.
\end{equation}
Please refer to TABLE~\ref{tab:notation} for the notations used throughout the paper.

\subsection{Context Graph Learning}
Once the graph is built and the features of input nodes are obtained, the next step is to promote individual features into a graph-level representation for robust group matching. To generate a robust group representation, it is necessary to take the following considerations into account: (1) Different group members have varying importance. For instance, occluded and outlier members are less important than members who are highly discriminative. (2) The group similarity is computed based on group pairs rather than individual groups. It is thus necessary to measure the mutual influence of individual group members of one group on the other. Therefore, it is highly important that the model be capable of understanding both intra- and inter- group context. To this end, we propose a novel GNN-based framework called Multi-Attenion Context Graph (MACG), where multiple attention mechanisms are proposed to discover and match discriminative features within and between groups. Specifically, MACG contains three attention modules for graph learning, which are detailed as follows.

\subsubsection{Intra-Group Attention}
To explore the context dependency within a group, graph neural networks allow nodes to encode and send messages to their neighbors through edge connections. The node representation is then updated accordingly by aggregating the received messages. Existing models have designed different strategies to aggregate messages, which can be generally summarized into two categories, \ie, pooling-based and attention based aggregators. For a graph node, pooling-based aggregators simply perform average/max/sum pooling on all the messages sent to the node. This operation is straightforward and has been widely adopted in various GNN models~\cite{DBLP:journals/corr/KipfW16,DBLP:conf/nips/DuvenaudMABHAA15,DBLP:conf/nips/HamiltonYL17}. However, pooling-based aggregators consider each message as being of equal importance, ignoring the correlations between nodes within the group. In our task, the correlations could be important cues for identifying groups. For example, occluded group members usually contain less discriminative information for identifying a group, thus the messages coming from those nodes are less important compared with other nodes. In contrast to pooling-based strategies, attention-based aggregators~\cite{DBLP:journals/corr/abs-1710-10903,DBLP:journals/corr/abs-1710-09599,DBLP:journals/corr/abs-1803-03735} calculate the importance weights of messages, and aggregate the messages by taking the weighted sums. We therefore develop an attention-based mechanism for intra-group message passing.

\begin{figure}[t]
    \centering
    \includegraphics[width=\linewidth]{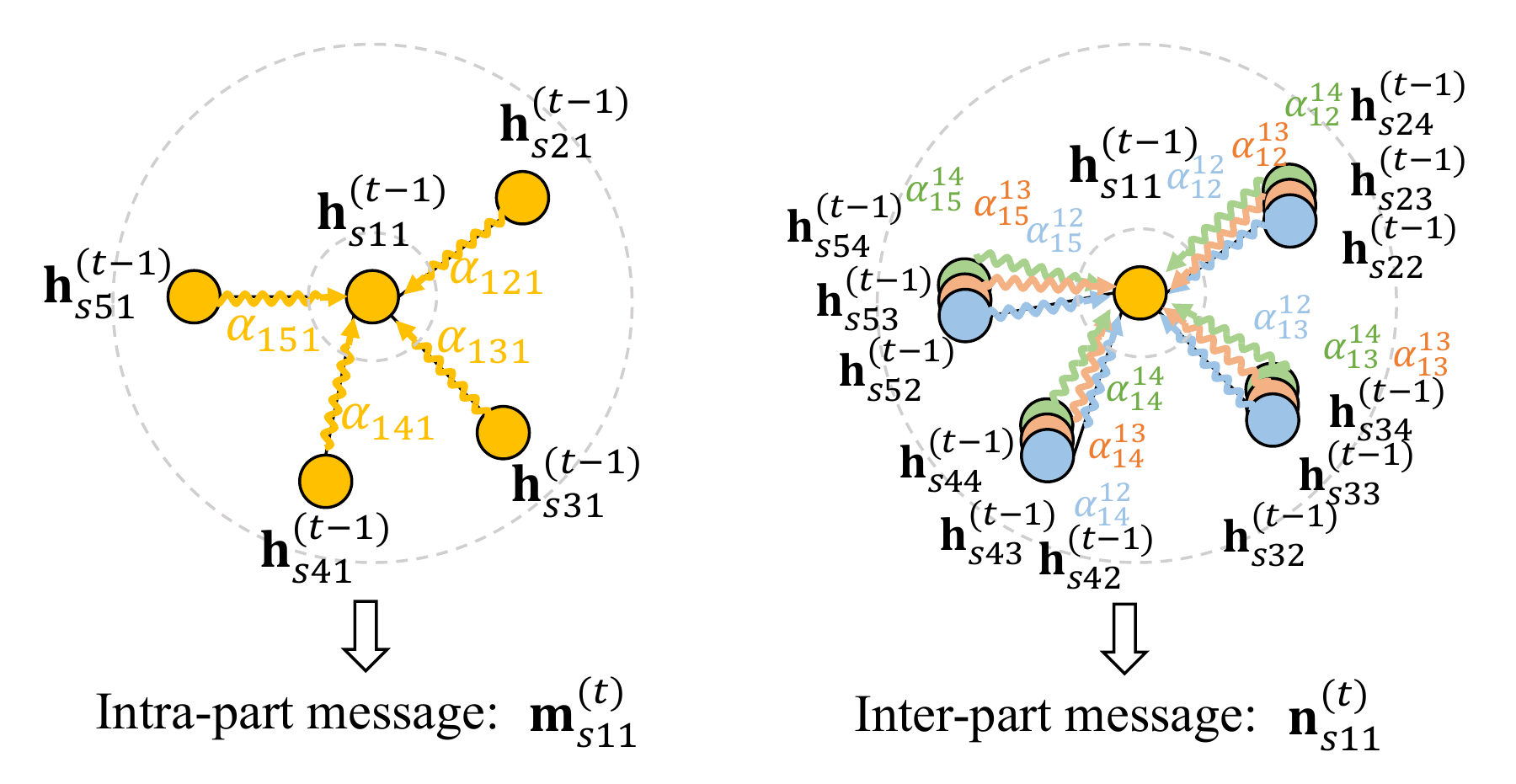}
    \caption{Illustration of intra- and inter-part attentions for a single node. Intra-part attention only receives messages from node features corresponding to the same body part, while inter-part attention receives messages from different part features. }
    \label{fig:partatt}
\end{figure}

Different from previous GNN models, where each node usually contains a single feature vector, every node in our context graph contains $P$ part-level features. Here, we introduce a part-level attention for each node, which not only considers the correlations between part-level features from the same part across different nodes, but also explores the dependencies of different parts from different nodes in case of misalignment or pose variations. In this way, the intra-graph attention is composed of two kinds of attention mechanisms, \ie, intra-part and inter-part attention.

For intra-part correlation computation, we consider a feature pair $(\mathbf{h}_{sip}^{(t-1)}, \mathbf{h}_{sjp}^{(t-1)})$ from the $s$-th group image $I_{s}$, where the features come from different nodes $(i, j)$ but correspond to the same person part $p$. The importance of the intra-part message sent from $j$ to $i$ can be calculated as follows:
\begin{equation}\label{eq:3}
    e_{ijp} = \phi(\textbf{W}_{e}^{(t-1)}\mathbf{h}_{sip}^{(t-1)}, \textbf{W}_{e}^{(t-1)}\mathbf{h}_{sjp}^{(t-1)}),
\end{equation}
where $\phi$ is a function that measures the correlation between the inputs, and $\mathbf{W}_{e}^{(t-1)}$ is a weight matrix that transforms the input features into higher-level representations. Following GAT~\cite{DBLP:journals/corr/abs-1710-10903}, we formulate $\phi$ as a fully connected layer based on the concatenation of the inputs. After obtaining all the importance weights $e_{ijp}, \ \forall j: (i,j)\in \mathcal{E}_s$, we calculate the attention weights by normalizing the importance weights with the softmax function: 
\begin{equation}\label{eq:4}
    \alpha_{ijp} = {\rm softmax}(e_{ijp}) = \frac{{\rm exp}(e_{ijp})}{\sum_{k}^{(i,k) \in \mathcal{E}_s } {\rm exp}(e_{ikp})}.
\end{equation}
Then, the intra-part messages passed to node $i$ can be aggregated by combining the neighbors' features with the corresponding attention weights:
\begin{equation}\label{eq:5}
    \mathbf{m}_{sip}^{(t)} = \sum_{j:(i,j) \in \mathcal{E}_s} {\alpha_{ijp} \textbf{W}_{e}^{(t-1)} \mathbf{h}_{sjp}^{(t-1)}}.
\end{equation}

Inter-part message passing can also be formulated with the attention mechanism similar to the intra-part attention, except that a node only receives messages from inter-part features. In this case, we consider a feature pair $(\mathbf{h}_{sip}^{(t-1)}, \mathbf{h}_{sjq}^{(t-1)})$, where the features come from different nodes $(i, j)$ and correspond to different person parts $(p,q)$. The importance of messages passed from $\mathbf{h}_{sjq}^{(t-1)}$ to $\mathbf{h}_{sip}^{(t-1)}$ is
\begin{equation}\label{eq:6}
    e_{ij}^{pq} = \phi(\textbf{W}_{e}^{(t-1)}\mathbf{h}_{sip}^{(t-1)}, \textbf{W}_{e}^{(t-1)}\mathbf{h}_{sjq}^{(t-1)}).
\end{equation}
Then the attention weights and inter-part message can be calculated respectively as follows:
\begin{equation}\label{eq:7}
    \alpha_{ij}^{pq} = {\rm softmax}(e_{ij}^{pq}) = \frac{{\rm exp}(e_{ij}^{pq})}{\sum_{k}^{(i,k) \in \mathcal{E}_s }  {\sum_{l}^{ l\neq p}}{\rm exp}(e_{ik}^{pl})},
\end{equation}
\begin{equation}\label{eq:8}
    \mathbf{n}_{sip}^{(t)} = \sum_{j}^{(i,j) \in \mathcal{E}_s} {\sum_{q}^{ q\neq p}} {\alpha_{ij}^{pq} \textbf{W}_{e}^{(t-1)} \mathbf{h}_{sjq}^{(t-1)}}.
\end{equation}

\begin{figure}[t]
    \centering
    \includegraphics[width=0.92\linewidth]{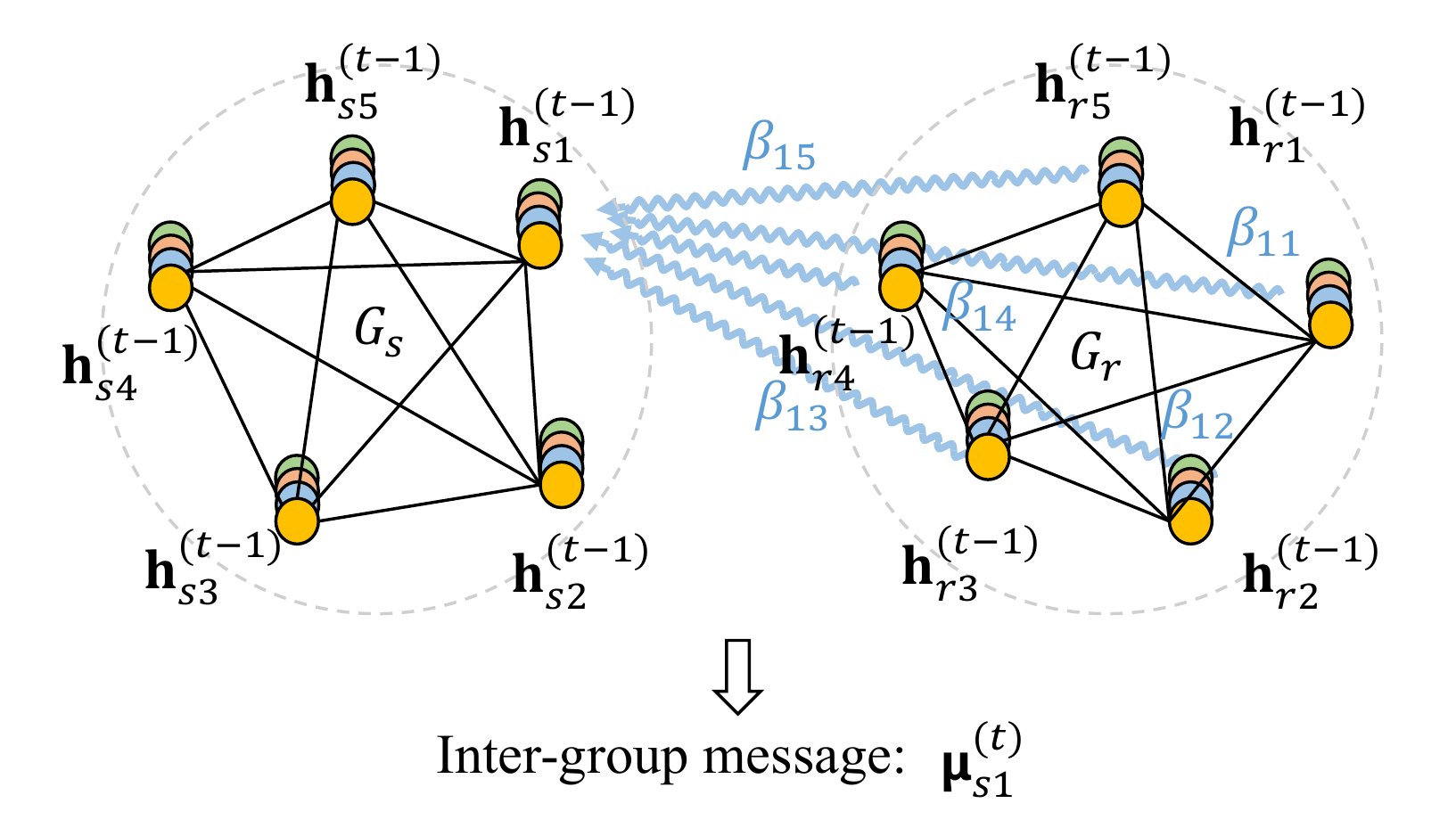}
    \caption{Illustration of inter-graph attention for a single node. We compute the node-level similarity between inter-graph nodes, and part-level features share the same set of attention weights. }
    \label{fig:interatt}
\end{figure}

After obtaining the intra-part message $\mathbf{m}_{sip}^{(t)}$ and the inter-part message $\mathbf{n}_{sip}^{(t)}$, they are further concatenated as the intra-graph message. Also note that we can employ multiple attention heads to calculate both the intra- and inter-part attentions, as in GAT~\cite{DBLP:journals/corr/abs-1710-10903}. The results obtained could then be concatenated or averaged for better representation. The intra-graph attention including both intra- and inter-part attention mechanisms is illustrated in Fig.~\ref{fig:partatt}.

\subsubsection{Inter-Group Attention}
Within-group context can be effectively modeled with the above-mentioned intra-group message passing. However, for group re-id, the objective is to calculate the similarity between group pairs. It is therefore necessary to explore the inter-group correlation. Intuitively, if a group pair shares the same ID, it is highly possible that there exists some correspondence between the individuals in the two groups. Meanwhile, a higher similarity for the individual pairs indicates a higher group-level similarity. Consider a node-level feature pair $(\mathbf{h}_{si}^{(t-1)}, \mathbf{h}_{rj}^{(t-1)})$ from two graphs $(\mathcal{G}_s,\mathcal{G}_r)$, where $i \in \mathcal{V}_s$ and $j \in \mathcal{V}_r$. We calculate the importance weights of inter-graph features as
\begin{equation}\label{eq:9}
    z_{ij} = \varphi(\mathbf{W}_{z}^{(t-1)}\mathbf{h}_{si}^{(t-1)}, \textbf{W}_{z}^{(t-1)}\mathbf{h}_{rj}^{(t-1)})),
\end{equation}
where $\varphi$ is simply an inner product layer since we only need to measure the similarity between graph nodes, and $\mathbf{W}_{z}^{(t-1)}$ is a projection matrix. Similarly, the inter-graph attention weights and the message passed from graph $\mathcal{G}_r$ to the $p$-th part of person $i$ in graph $\mathcal{G}_s$ can be calculated respectively as follows:
\begin{equation}\label{eq:10}
    \beta_{ij} = {\rm softmax}(z_{ij}) = \frac{{\rm exp}(z_{ij})}{\sum_{k\in \mathcal{V}_r} {\rm exp}(z_{ik})},
\end{equation}
\begin{equation}\label{eq:11}
    \bm{\mu}_{sip}^{(t)} = \sum_{j \in \mathcal{V}_r}\sum_{q} {\beta_{ij} \textbf{W}_{z}^{(t-1)} \mathbf{h}_{rjq}^{(t-1)}}.
\end{equation}
Note that here we consider person-level similarity for inter-graph attention calculations. Therefore, the part-level features in a node share the same set of inter-graph attention weights (\ie, $\beta_{ij}$). Fig.~\ref{fig:interatt} illustrates the intra-graph attention mechanism.

After obtaining the intra-graph (including intra- and inter-part) and inter-graph messages, the node features are updated with fully connected layers by concatenating previous features and all three types of messages:
\begin{equation}\label{eq:12}
    \mathbf{h}_{sip}^{(t)} = {\rm MLP}(\mathbf{h}_{sip}^{(t-1)}, \mathbf{m}_{sip}^{(t)}, \mathbf{n}_{sip}^{(t)}, \bm{\mu}_{sip}^{(t)}) .
\end{equation}

\begin{algorithm}[t]
\caption{Multi-Attention Context Graph Learning}
\label{alg:1}
\begin{algorithmic}[1]
\REQUIRE Cropped images $\{{I}_{si}, I_{rj}\} _{i \in 1,...,N_s, j\in 1,...,N_r}$ from a pair of group images $({I}_{s}, I_{r})$ , group label $y_{g}$, person label $y_p$, the ground-truth permutation matrix $\mathbf{S}^{gt}$\\
\ENSURE Multi-Attention  Context  Graph (MACG) Network
\WHILE{not converge}
\STATE Extract features $\{\mathbf{h}_{sip}^{(0)}, \mathbf{h}_{rjp}^{(0)}\} \leftarrow \{{I}_{si}, I_{rj}\}$ by Eq.~(\ref{eq:1})
\STATE Build the context graphs $\mathcal{G}_s, \mathcal{G}_r$ by Eq.~(\ref{eq:2})
\FOR{$t \leftarrow 1,...,T$}
\STATE Calculate the intra-part messages by Eqs.~(\ref{eq:3})-(\ref{eq:5}): \\
$\mathbf{m}_{sip}^{(t)} \leftarrow \{ \mathbf{h}_{sjp}^{(t-1)}\} _{j \in \mathcal{V}_s}$,
$\mathbf{m}_{rip}^{(t)} \leftarrow \{ \mathbf{h}_{rjp}^{(t-1)}\} _{j \in \mathcal{V}_r}$ \\
\STATE Calculate the inter-part messages by Eqs.~(\ref{eq:6})-(\ref{eq:8}) \\
$\mathbf{n}_{sip}^{(t)} \leftarrow \{ \mathbf{h}_{sjq}^{(t-1)}\} _{j \in \mathcal{V}_s, q \neq p}$, $\mathbf{n}_{rip}^{(t)} \leftarrow \{ \mathbf{h}_{rjq}^{(t-1)}\} _{j \in \mathcal{V}_r, q \neq p}$ \\
\STATE Calculate the inter-group messages by Eqs.~(\ref{eq:9})-(\ref{eq:11}) \\
$\bm{\mu}_{sip}^{(t)} \leftarrow \{ \mathbf{h}_{rjq}^{(t-1)}\} _{j \in \mathcal{V}_s}$,  $\bm{\mu}_{rip}^{(t)} \leftarrow \{ \mathbf{h}_{sjq}^{(t-1)}\} _{j \in \mathcal{V}_r}$ \\
\STATE Update the node representations by Eq.~(\ref{eq:12}) \\
$\mathbf{h}_{sip}^{(t)} \leftarrow (\mathbf{h}_{sip}^{(t-1)}, \mathbf{m}_{sip}^{(t)}, \mathbf{n}_{sip}^{(t)}, \bm{\mu}_{sip}^{(t)})$ \\
$\mathbf{h}_{rip}^{(t)} \leftarrow (\mathbf{h}_{rip}^{(t-1)}, \mathbf{m}_{rip}^{(t)}, \mathbf{n}_{rip}^{(t)}, \bm{\mu}_{rip}^{(t)})$ \\
\ENDFOR

\STATE Extract the graph-level feature by Eqs.~(\ref{eq:13})-(\ref{eq:15}) \\
${\mathbf{h}}_{s},\mathbf{h}_{r} \leftarrow \{{\mathbf{h}}_{si}^{(T)}, {\mathbf{h}}_{rj}^{(T)}\} _{i \in \mathcal{V}_s, j \in \mathcal{V}_r}$
\STATE Calculate the losses by Eqs.~(\ref{eq:16})-(\ref{eq:21}) \\
$L_{g}^{pair}, L_{p}^{pair}, L_{pce} \leftarrow \{\mathbf{h}_{s},\mathbf{h}_{r}, {\mathbf{h}}_{si}^{(T)}, {\mathbf{h}}_{rj}^{(T)}\} _{i \in \mathcal{V}_s, j \in \mathcal{V}_r}$
\STATE Back-propagation and update the MACG Network
\ENDWHILE
\end{algorithmic}
\end{algorithm}

\subsubsection{Correspondence Learning}
The above feature updating steps using intra- and inter-graph attention mechanisms are repeated for $T$ rounds and then the model is designed to learn group and person correspondence, respectively.
We first construct a graph-level representation via the \emph{readout} operation. Here, we simply apply self-attention on graph nodes and the final graph representation ${\textbf{h}}_{s}$ is a weighted sum of the node-level features:
\begin{equation}\label{eq:13}
    u_{i} = \textbf{W}_{u}^{(T)}\mathbf{h}_{si}^{(T)},
\end{equation}
\begin{equation}\label{eq:14}
    \gamma_{i} = \frac{{\rm exp}(u_{i})}{\sum_{k\in \mathcal{V}_s} {\rm exp}(u_{k})},
\end{equation}
\begin{equation}\label{eq:15}
    {\mathbf{h}}_{s} = \sum_{i \in \mathcal{V}_s} {\gamma_{i} \textbf{W}_{u}^{(T)} {\textbf{h}}_{si}^{(T)}}.
\end{equation}
We can also calculate ${\textbf{h}}_{r}$ for graph $\mathcal{G}_r$ in a similar manner.

\begin{figure}[t]
    \centering
    \includegraphics[width=\linewidth]{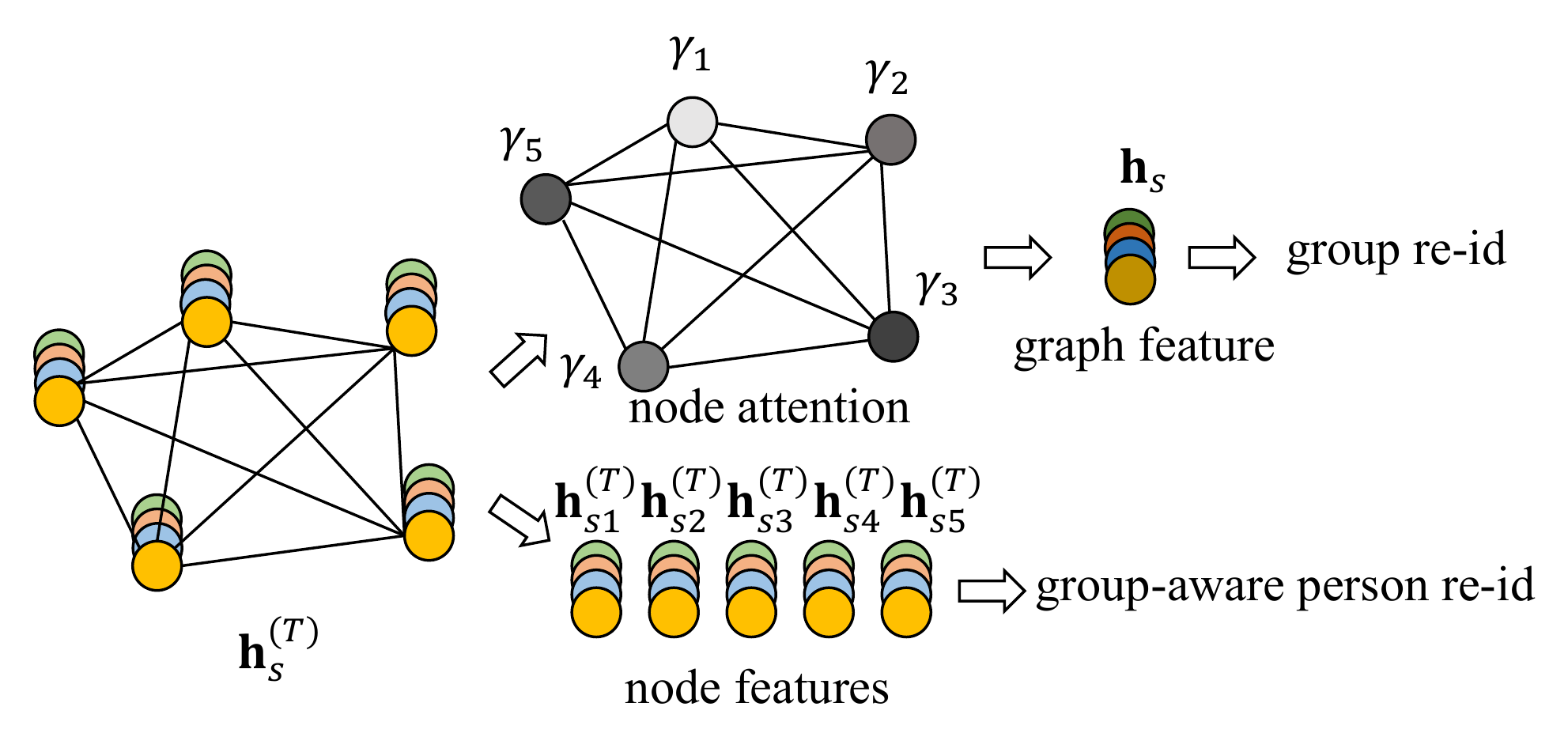}
    \caption{Illustration of the inference stages for group re-id and group-aware person re-id.}
    \label{fig:inference}
\end{figure}

To learn group correspondence, the pairwise loss function is adopted to pull features of the same group close together and push different groups far apart:
\begin{equation}\label{eq:16}
    L_{g}^{pair} = \{0, m - y_{g}(1-\|{\mathbf{h}}_{s} - {\mathbf{h}}_{r} \|^2)\},
\end{equation}
where $y_{g}$ is the label of the pair, and $m$ is the margin. $y_{g}=1$ when the pair shares the same group ID and $y_{g}=-1$ when the pair consists of different groups. Note that, similar to Graph Matching Net~\cite{DBLP:conf/icml/LiGDVK19}, a triplet loss can also be applied for training the context graph.

In addition to considering global group-level representations, local person-level information is also important for constructing global correspondence. For person-level correspondence learning, we also employ a pair-wise loss function:
\begin{equation}\label{eq:17}
    L_{p}^{pair} = \sum_{i \in \mathcal{V}_s, j \in \mathcal{V}_r}\sum_{p}{\rm max}\{0, m - y_{p}^{ij}(1 - \|{\mathbf{h}}_{sip} - {\mathbf{h}}_{rjp} \|^2)\},
\end{equation}
where $y_{p}^{ij}$ indicates whether the $i$-th node in $\mathcal{V}_s$ and the $j$-th node in $\mathcal{V}_r$ belong to the same identity.
In addition, we employ a permutation cross-entropy loss~\cite{DBLP:journals/corr/abs-1904-00597} to learn better person-level correspondence. Specifically, an affinity matrix $\mathbf{M}$ is first computed to measure the node-level affinity between two graphs:
\begin{equation}\label{eq:18}
    \mathbf{M}_{i,j} = {\rm exp}(\frac{\mathbf{h}_{si}^{\mathsf{T}}\mathbf{A}\mathbf{h}_{rj}}{\tau}),
\end{equation}
where $\mathbf{A}$ is a weight matrix to learn the affinity function, and $\tau$ is a hyper-parameter. A Sinkhorn operator \cite{sinkhorn1964} is then applied to $\mathbf{M}$ to generate the permutation matrix:
\begin{equation}\label{eq:19}
    \mathbf{S} = {\rm Sinkhorn}(\mathbf{M}),
\end{equation}
where the Sinkhorn operator iteratively performs row-wise and column-wise normalization on the input matrix until convergence. Finally, the cross entropy loss is applied between the predicted permutation matrix $\mathbf{S}$ and the ground-truth matrix $\mathbf{S}^{gt}$:
\begin{equation}\label{eq:20}
    L_{pce} = - \sum_{i \in \mathcal{V}_s, j \in \mathcal{V}_r}(\mathbf{S}_{i,j}^{gt} {\rm log}\mathbf{S}_{i,j} + (1 - \mathbf{S}_{i,j}^{gt}){\rm log}(1 - \mathbf{S}_{i,j})),
\end{equation}
where $\mathbf{S}^{gt}\in \mathcal{R}^{N_s\times N_r}$ is a binary matrix, and $\mathbf{S}^{gt}_{i,j}=1$ if the $i$-the person in $\mathcal{G}_s$ and the $j$-th person in $\mathcal{G}_r$ belong to the same identity.

During training, the overall loss $L$ is a linear combination of all the above loss functions:
\begin{equation}\label{eq:21}
    L = L_{g}^{pair} + L_{p}^{pair} + L_{pce}.
\end{equation}
We summarize the training process of our framework in Algorithm~\ref{alg:1}.

\subsection{Model Inference}
After the model is trained, we can perform two types of group-based re-id tasks, as illustrated in Fig.~\ref{fig:inference}. Specifically, the graph-level representations from the readout attention module are directly employed for group re-id. For group-aware person re-id, the node-level features already contain discriminative context information as they receive messages from both intra-group and inter-group members. In addition, we can also utilize the person correspondence learning module to further reduce the ambiguity between people with similar appearances. The inference of group re-id and group-aware person re-id can be jointly computed in our framework. The results of group re-id and group-aware person re-id are discussed in Sec.~\ref{sec:group} and Sec.~\ref{sec:person}, respectively.

\section{CUHK-SYSU-Group Dataset}\label{sec:data}
In addition to the proposed MACG model, another important contribution of this paper is the newly collected CUHK-SYSU-Group (CSG) dataset. To the best of our knowledge, CSG is the largest dataset for group re-identification, consisting of more than 3.8$K$ images from 1.5$K$ labeled groups. In the following, we will introduce the detailed collection process and the statistical distribution of this dataset. It is worth noting that our CSG dataset is suitable for evaluating both group re-id and group-aware single person re-id methods, owing to the instance-level annotation.

\subsection{Dataset Collection}
Creating a large-scale group re-id dataset is very challenging. First of all, there is no strict guideline of how to define a group. People that are geometrically close to each other can often be considered as a group, and existing datasets typically follow this criterion. However, we argue that sometimes group members are not necessarily spatially close. Please refer to the example in Fig.~\ref{subfig-1-3}, where we can observe interactions between group members, even if they are far away from each other. In this situation, we still consider them to be a group. Therefore, during annotation we manually determine a group by jointly considering the spatial and interactive cues. Secondly, data collection and group annotation are both labor-intensive. This is probably why the size of existing datasets are rather limited. In this work, we construct the group re-id dataset based on the CUHK-SYSU dataset, which was originally designed for the person search task. Thanks to the person-level IDs and bounding box annotations in CUHK-SYSU, we can take the advantage of them for group annotation. Specifically, we first discover the people that simultaneously appear in different images as group candidates. Then we manually and empirically select the groups according to the spatial layout and interactions between people. Following previous protocol~\cite{xiao2018group}, we annotate the groups with the same ID when they have more than 60\% of members in common. We further consolidate the dataset by refining group annotations with the help of ten volunteers. Specifically, after the groups are annotated, we only keep the groups when at least seven volunteers agree on the correctness of the annotations.

\begin{table}[t]
    \centering
    \caption{Statistical comparisons between CSG and existing group re-id datasets. MCTS denotes i-LIDS MCTS dataset, RG denotes Road Group dataset and DG denotes DukeMTMC Group dataset.}
    \label{tab:dataset}
    \begin{tabular}{c|cccc}
    \toprule[1pt]
        Datasets & MCTS~\cite{DBLP:conf/bmvc/ZhengGX09} & RG~\cite{xiao2018group} & DG~\cite{xiao2018group} & \textbf{CSG} \\ \hline
        \# Image & 274 & 324 & 354 & \textbf{3839} \\
        \# Group & 64 & 162 & 177  & \textbf{1558}\\
        \# Viewpoint & 8& 2& 8 & \textbf{diversified}\\
    \bottomrule[1pt]
    \end{tabular}
\end{table}

\begin{figure}[t]
    \centering
    \includegraphics[width=\linewidth]{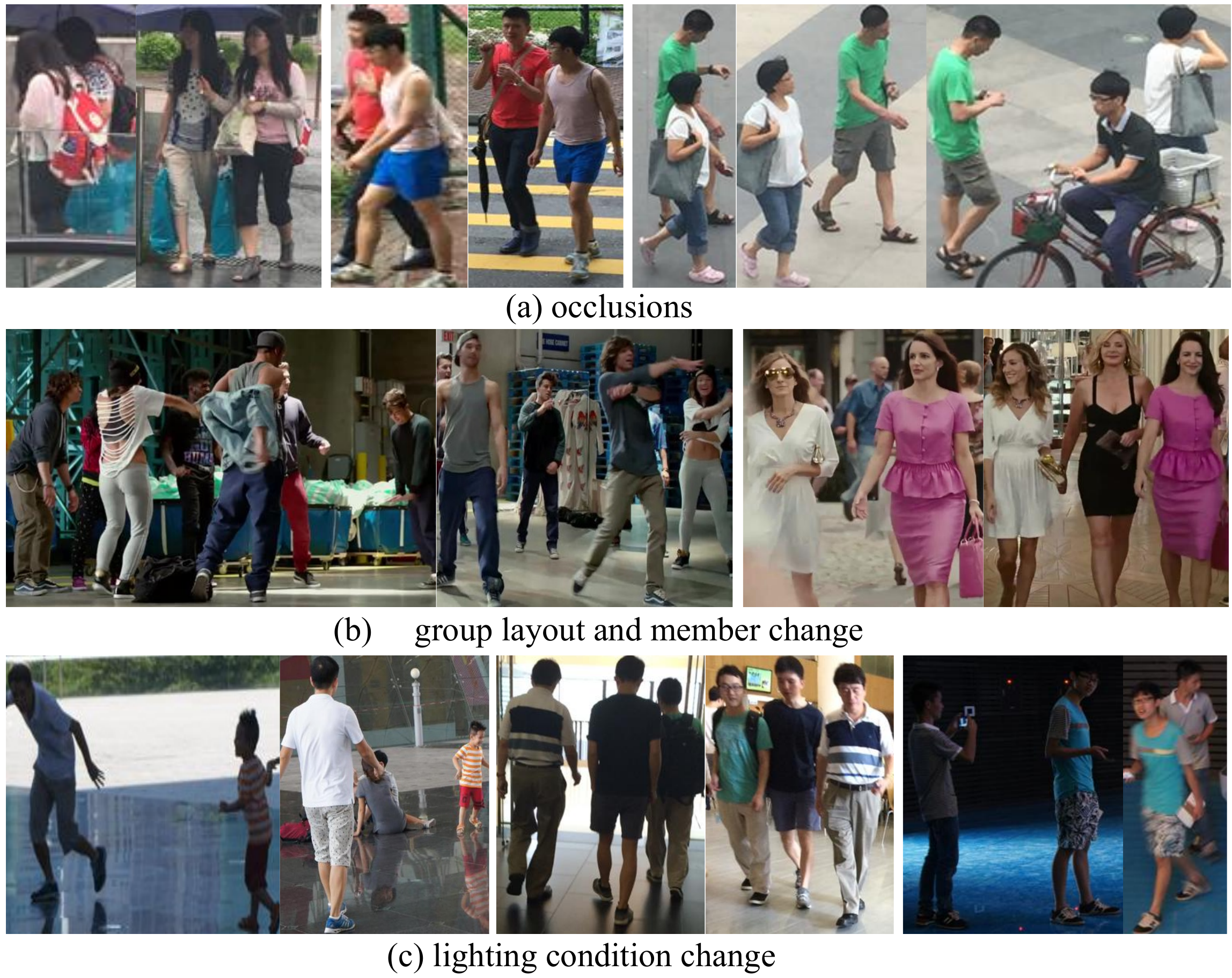}
    \vspace{-4mm}
    \caption{Visualization of some examples from the CUHK-SYSU-Group dataset. This dataset contains various challenging situations, such as occlusions, group layout and member changes, lighting condition changes, etc.}
    \label{fig:dataset}
\end{figure}

\begin{table*}[t]
\centering
\caption{Comparison with the state-of-the-art group re-id methods. R-$k$ ($k$=1, 5, 10, 20) denotes the Rank-$k$ accuracy (\%).}
\label{tab:cmp1}
\renewcommand{\arraystretch}{1.5}
\begin{tabular*}{\textwidth}{@{\extracolsep{\fill}} c|cccc|cccc|cccc|cccc}
\hline
         & \multicolumn{4}{c|}{CSG} & \multicolumn{4}{c|}{i-LIDS MCTS} & \multicolumn{4}{c|}{DukeMTMC Group} & \multicolumn{4}{c}{Road Group} \\ 
         & R-1  & R-5 & R-10 & R-20 & R-1   & R-5   & R-10   & R-20   & R-1  & R-5 & R-10 & R-20 & R-1   & R-5   & R-10   & R-20  \\ \hline \hline
CRRRO-BRO~\cite{DBLP:conf/bmvc/ZhengGX09}  & 10.4   & 25.8  & 37.5  & 51.2  & 23.3 & 54.0 & 69.8 & 82.7 & 9.9   & 26.1  & 40.2  & 64.9    & 17.8  & 34.6  & 48.1   & 62.2      \\
Covariance~\cite{DBLP:conf/icpr/CaiTP10}    & 16.5   & 34.1  & 47.9  & 67.0     & 26.5 & 52.5 & 66.0 & 90.9  & 21.3  & 43.6  & 60.4  & 78.2    & 38.0    & 61.0    & 73.1   & 82.5 \\ 
PREF~\cite{DBLP:conf/iccv/LisantiMBF17}       &19.2   & 36.4  & 51.8  & 70.7     & 30.6  & 55.3 & 67.0  & 92.6   & 22.3  & 44.3  & 58.5  & 74.4    & 43.0    & 68.7  & 77.9   & 85.2 \\
BSC+CM~\cite{DBLP:conf/icip/ZhuCY16}    & 24.6   & 38.5  &55.1  & 73.8     & 32.0  & 59.1 & 72.3  & 93.1  & 23.1  & 44.3  & 56.4  & 70.4    & 58.6  & 80.6  & 87.4   & 92.1 \\
MGR~\cite{xiao2018group}       & 57.8   & 71.6  & 76.5  & 82.3    & 38.8  & 65.7  & 82.5   & 98.8 & 48.4  & 75.2  & 89.9  & \textbf{94.4}    & 80.2  & 93.8  & 96.3   & 97.5       \\ \hline
\textbf{MACG}   &\textbf{63.2}   & \textbf{75.4}  & \textbf{79.7} & \textbf{84.4}     & \textbf{45.1}   & \textbf{70.4}  & \textbf{84.9} & \textbf{99.1} &\textbf{57.4}   & \textbf{79.0}  & \textbf{90.3} & 94.3   & \textbf{84.5}  & \textbf{95.0}  & \textbf{96.9}   & \textbf{98.1}\\ \hline
\end{tabular*}
\end{table*}

\subsection{Dataset Statistics}
Our CSG dataset contains 3,839 images of 1,558 groups, with about 3.5$K$ annotated people and 9$K$ bounding boxes. Each group contains at least two people and the largest group contains eight people. The average number of group members on CSG is 2.5. We statistically compare CSG with three existing group re-id datasets, \ie, i-LIDS MCTS~\cite{DBLP:conf/bmvc/ZhengGX09}, Road Group~\cite{xiao2018group} and DukeMTMC Group~\cite{DBLP:conf/eccv/RistaniSZCT16,xiao2018group}, as listed in TABLE~\ref{tab:dataset}. We can see that CSG is an order of magnitude larger than existing datasets with respect to the number of images and groups. Additionally, the viewpoints and data sources of CSG are more diverse than in existing datasets. For example, existing datasets only contain images from surveillance cameras with fixed camera viewpoints. In contrast, in CSG (as with CUHK-SYSU), images are captured from hand-held cameras as well as movie snapshots, which have much more diversified viewpoints, backgrounds and lighting conditions. 
Some examples from CSG are illustrated in Fig.~\ref{fig:dataset}. We can observe from the examples that this dataset contains various challenging scenarios, including occlusions, group layout and member changes, and lighting condition changes, and these challenges may even co-exist in a single group. On CSG, about 26\% of the groups contain persons that are occluded, while only 4\% of the images are captured from exactly the same viewpoint, making most group images contain layout/member or lighting condition changes. All the above factors make the CSG dataset more challenging and suitable for the task of group re-id.

\section{Experimental Results}\label{sec:experiments}
In this section, we evaluate our model on its ability to associate groups and group members. We conduct extensive experiments on our CSG dataset and three existing group re-id datasets. We first compare our model with the state-of-the-art methods, and then evaluate our method on group-aware person re-id, by directly employing node features. We also conduct a comprehensive ablation study to verify the contribution of each model component, as well as the sensitivity of the proposed framework under different settings and parameters. We further visualize the attention weights to better understand the attention mechanisms. Finally, we analyze the impacts of pedestrian and group detectors.

\subsection{Implementation Details and Experimental Setup}
We employ ResNet50~\cite{DBLP:conf/cvpr/HeZRS16} pretrained on ImageNet~\cite{DBLP:conf/cvpr/DengDSLL009} as the backbone. The person images are resized to 256$\times$128 as inputs. The initial learning rate is set to 0.0003 and is reduced by a factor of 10 at the 80-th and 160-th epochs, with the training stage terminating at the 200-th epoch. As groups are of different sizes, we construct graphs with equal numbers of nodes to facilitate implementation, and add dummy nodes to the groups with limited members. We only perform person correspondence learning on positive group pairs, as there exists no correspondence for negative pairs. We use two-layer (\ie, $T=2$) GNNs in our framework. We train our model on one Tesla V100 GPU and it takes about 28 hours for the model to converge on the CSG dataset.

We manually split CSG into fixed training and test sets, where 859 groups are utilized for training and the remaining 699 groups for testing. We also ensure that the groups in the test set are captured from different viewpoints. During testing, the images in the test set are sequentially selected as the probe, while all the remaining images are regarded as the gallery set. In this way, there is no overlap of viewpoints between the probe and gallery images. Additionally, we add 5$K$ group images containing 20$K$ person bounding boxes as distractors in the gallery set, making it comparable in scale to traditional person re-id datasets. As for iLID-MCTS, Duke Group, and Road Group, we randomly partition the datasets into training sets and test sets with equal sizes. The final result is obtained by averaging the results of 10 random splits.
We use the Cumulative Matching Characteristics (CMC) as the evaluation metric.

\subsection{Group Re-Identification}\label{sec:group}
To evaluate the group re-id performance, we compare our full model with a few state-of-the-art methods specifically designed for group re-id: CRRRO-BRO~\cite{DBLP:conf/bmvc/ZhengGX09}, Covariance~\cite{DBLP:conf/icpr/CaiTP10}, PREF~\cite{DBLP:conf/iccv/LisantiMBF17}, BSC+CM~\cite{DBLP:conf/icip/ZhuCY16} and MGR~\cite{xiao2018group}. The quantitative results are illustrated in TABLE~\ref{tab:cmp1}. 

We find that the performance of most previous methods is limited, mainly due to the following two reasons. Firstly, most previous methods employ hand-crafted features for group representation. Secondly, these methods merely consider the global representation of the entire group, ignoring rich context information from group members. For example, CRRRO-BRO~\cite{DBLP:conf/bmvc/ZhengGX09} designs a center rectangular ring ration-occurrence descriptor (CRRRO), which tries to find a stable representation against a relative position change between two people, as well as a block-based ratio-occurrence descriptor (BRO) for non-center-rotation changes. However, associating groups of people is much more difficult than modeling the relative positions of person pairs. This is why CRRRO-BRO~\cite{DBLP:conf/bmvc/ZhengGX09} achieves a relatively better performance on datasets with smaller group sizes (\eg, i-LIDS MCTS, in which most groups only contain two people). The covariance~\cite{DBLP:conf/icpr/CaiTP10} descriptor measures the distance between two groups as the dissimilarity between the covariance matrices. However, the covariance matrices are calculated based on local pixel values, which can be easily influenced by the background noise that commonly appears in group scenes. Therefore, its performance is limited. PREF~\cite{DBLP:conf/iccv/LisantiMBF17} learns a feature dictionary for single person re-id and then transfers it for group appearance encoding. In real scenarios, group appearance is more variable than single person appearance and thus the performance of such an unsupervised method is also limited. BSC+CM~\cite{DBLP:conf/icip/ZhuCY16} explores patch correspondence between group images, but its performance is still limited due to the complex dynamics within a group.

\begin{figure}[t]
    \centering
    \includegraphics[width=\linewidth]{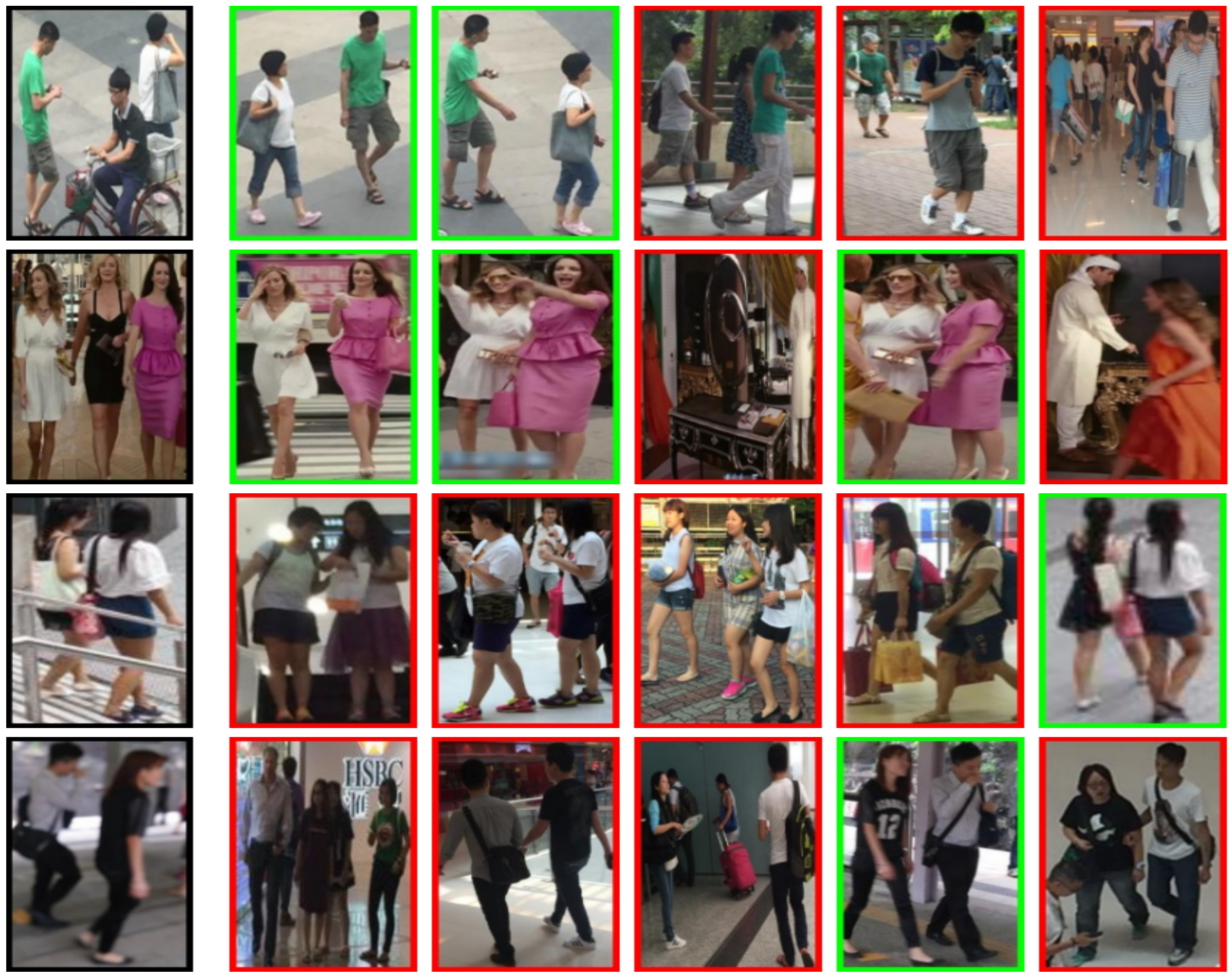}
    \caption{Visualization of group re-id results. The first image is the query, whilst the rest are the Rank-1 to Rank-5 (from left to right) retrieved results. The green and red bounding boxes denote correct and incorrect matches, respectively.}
    \label{fig:vis_group}
    \vspace{-4mm}
\end{figure}

Among existing methods, MGR~\cite{xiao2018group} is the only one that employs deep models for feature representation. More importantly, MGR explores the multi-granular correspondence between groups using a graph model and it achieves evidently better performance than previous models. The proposed multi-attention context graph differs from MGR in the following aspects. First of all, MGR~\cite{xiao2018group} focuses on learning redundant multi-granular sub-group correspondence to improve robustness of re-identification. The improvement from the multi-granular representation is more significant for groups with large numbers of members, and it is not as obvious for datasets containing less people (\eg, i-LIDS MCTS). However, there are often higher chances to observe small groups than large groups in real scenarios. In contrast, for the proposed multi-attention framework, there is no explicit constraint on multi-granular information and we can clearly observe consistent improvement on all the datasets. Second, MGR~\cite{xiao2018group} contains three stages that are optimized separately, whilst the proposed framework is built upon graph neural networks and is end-to-end learnable. In addition, the proposed framework outperforms all existing state-of-the-art models, demonstrating the effectiveness of the proposed MACG model.

We visualize some group re-id results in Fig.~\ref{fig:vis_group}, where the first two examples illustrate the cases that group member changes between query and gallery images. It can be observed that the correct matches can be successfully retrieved in these cases, even if there exist occlusions in the first example. The last two examples show the failure cases, where the gallery groups usually contain persons sharing similar appearance with both pedestrians in the query set. In this case, it would be more difficult for the model to retrieve the correct match.

\begin{figure*}[t]
    \centering
    \includegraphics[width=0.48\linewidth]{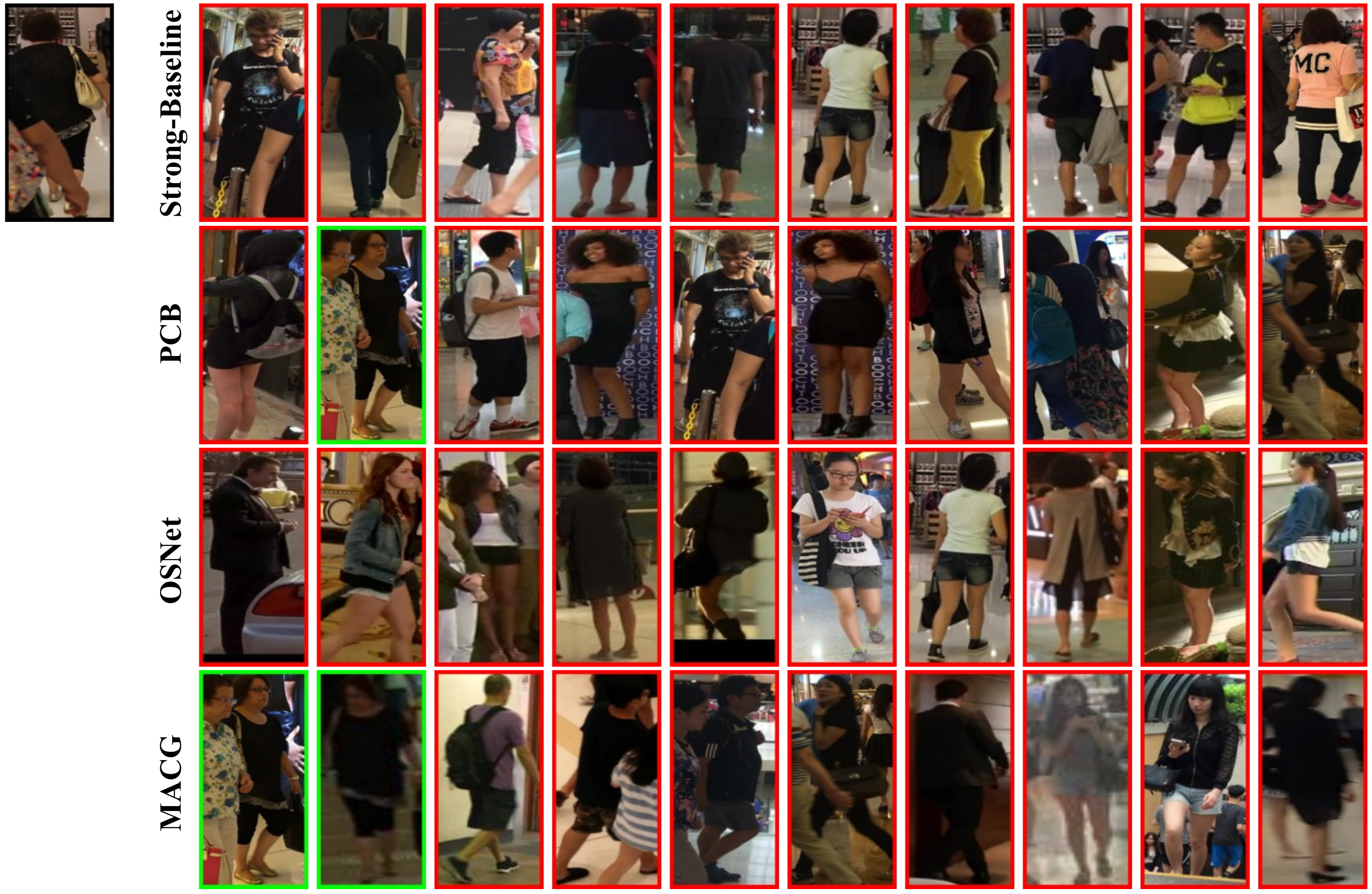}
    \hspace{-1mm}
    \includegraphics[width=0.48\linewidth]{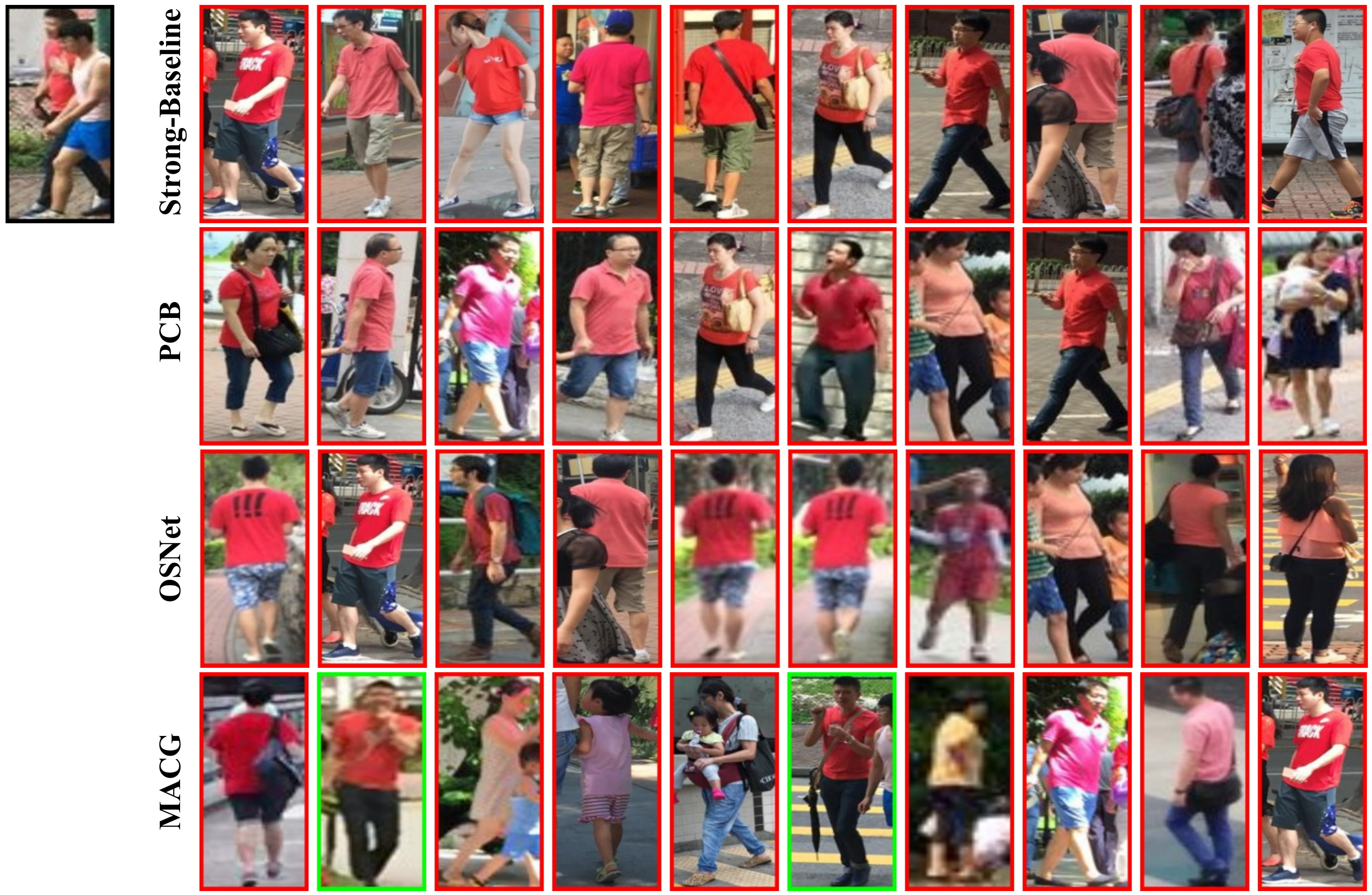}
    \includegraphics[width=0.48\linewidth]{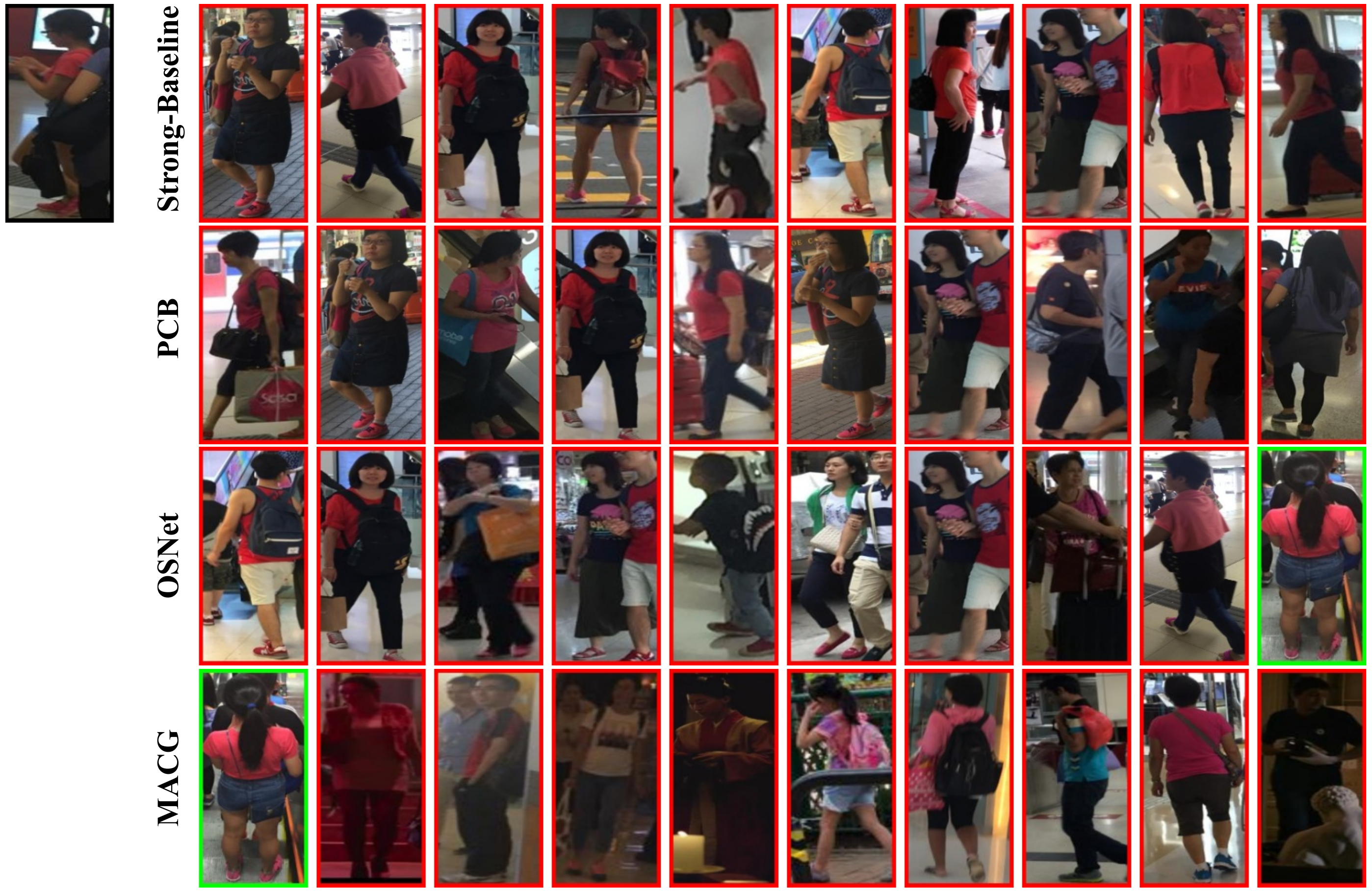}
    \hspace{-1mm}
    \includegraphics[width=0.48\linewidth]{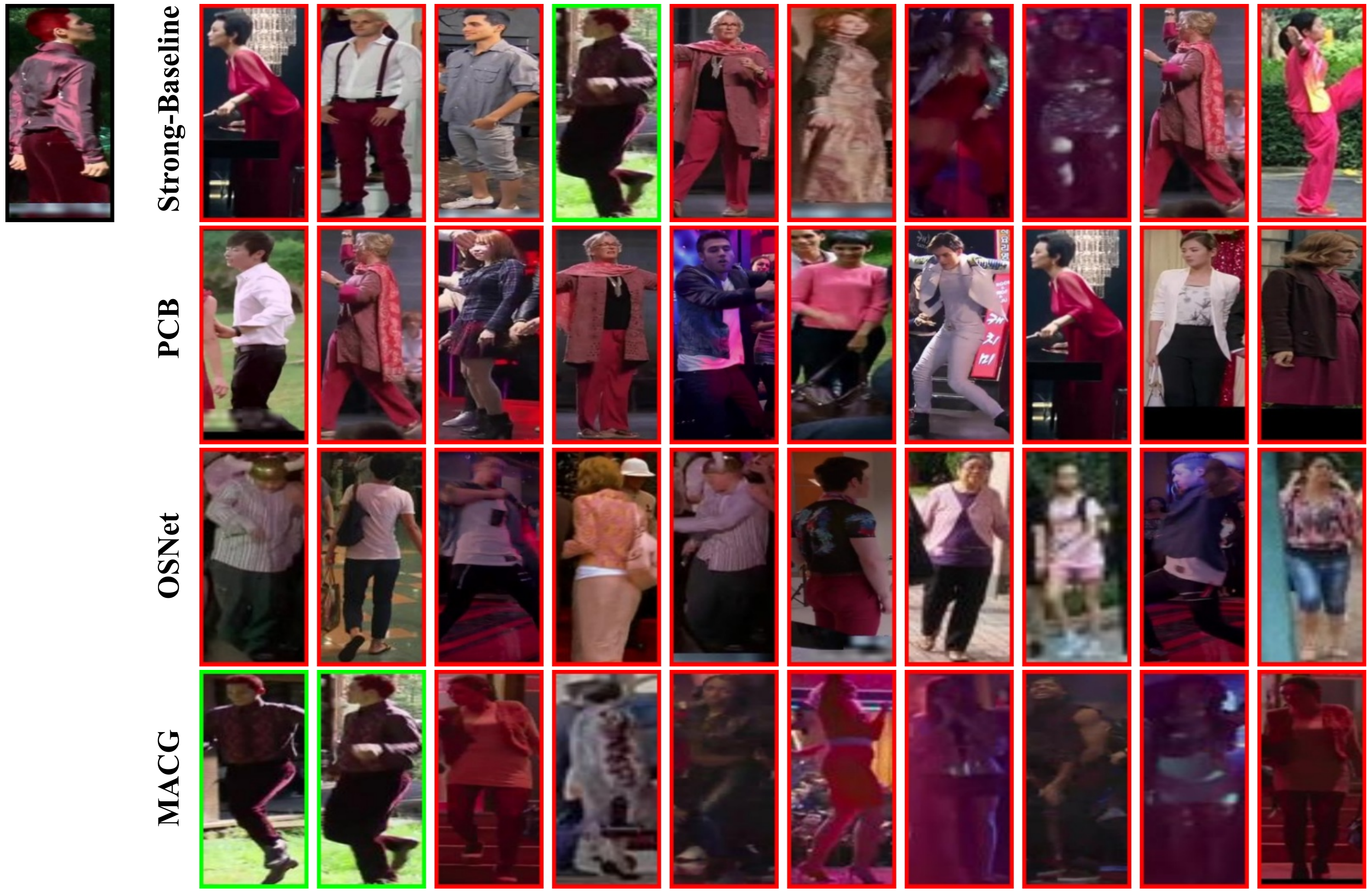}
    \caption{Visualization of re-id results using single person re-id models and the proposed MACG model. The first image is the query, whilst the rest are the Rank-1 to Rank-10 (from left to right) retrieved results. The green and red bounding boxes denote correct and incorrect matches, respectively.}
    \label{fig:vis}
\end{figure*}

\subsection{Group-Aware Person Re-Identification}\label{sec:person}
Group context information has proven to be beneficial for reducing ambiguity in single person re-id~\cite{DBLP:conf/bmvc/ZhengGX09,xiao2018group,Yan_2019_CVPR}. In this work, we explore both group correspondence learning and person correspondence learning, which could be naturally utilized for single person re-id under the guidance of group context. Specifically, we utilize the node-level features to measure the similarity of person pairs between groups. 


\begin{table}[t]
\centering
\caption{Comparative results for single person re-id with group context information.}
\label{tab:person}
\renewcommand{\arraystretch}{1.5}
\begin{tabular}{l|c|ccccc} \hline
Model & Group & mAP & R-1 & R-5 & R-10 & R-20\\ \hline \hline
  Our CNN & - & 61.3 & 61.3 & 75.2& 81.0& 86.5\\ 
  Strong-Baseline\cite{8930088}  & - & 63.4 & 63.1& 80.3& 84.3& 87.1\\ 
  PCB\cite{DBLP:conf/eccv/SunZYTW18} &- & 60.5& 63.7 &80.2& 83.9& 85.8\\ 
  OSNet\cite{DBLP:conf/iccv/ZhouYCX19} &- & 61.3& 62.4 &77.0& 81.0& 84.6\\ \hline
  CG \cite{Yan_2019_CVPR}& $\surd$  & 62.1 & 62.7& 78.4& 82.6& 87.2\\ 
  MGR~\cite{xiao2018group}  & $\surd$  & 63.3 & 63.8& 79.9& 83.8& 87.4\\  \hline
  MACG w/o CL & $\surd$  & 64.2 & 63.9   & 79.7  & 83.4 & 87.5 \\
  \textbf{MACG} & $\surd$  & \textbf{66.5} & \textbf{65.6}   & \textbf{80.5}  & \textbf{84.6} & \textbf{88.1} \\ \hline
\end{tabular}
\end{table}

We compare our method with the following two types of models for person re-id: 1) traditional person re-id models without group information, \ie, our baseline CNN model, Strong-Baseline~\cite{8930088}, PCB\cite{DBLP:conf/eccv/SunZYTW18}, and OSNet\cite{DBLP:conf/iccv/ZhouYCX19}.
2) MGR~\cite{xiao2018group} that explores group correspondence learning to help single person re-id, and Context Graph (CG)~\cite{Yan_2019_CVPR} which also explores group context for person re-id. The results are reported in TABLE~\ref{tab:person}. We find that the models using group information generally achieve better performance than the baseline person re-id models, which further validates the effectiveness and benefits of exploiting group context for person re-id. On the one hand, compared with our baseline CNN, the proposed MACG improves the Rank-1 matching accuracy and mAP by 4.3\% and 5.2\%, respectively. On the other hand, MACG also achieves notable improvement compared with recent state-of-the-art person re-id models. 
We also observe that, compared with MGR~\cite{xiao2018group} and CG~\cite{Yan_2019_CVPR}, the proposed model achieves further improvement. Compared with MGR~\cite{xiao2018group}, our framework applies a multi-level attention mechanism to better facilitate context message passing among groups. Although CG~\cite{Yan_2019_CVPR} employs group context, it only considers instance-level similarity learning in the loss function. In contrast, the proposed framework jointly optimizes group and instance correspondence learning, thus yielding better performance. If our model is trained without the person correspondence loss (\ie, MACG w/o CL), the improvement is less significant. The above results demonstrate that group context information is not only useful for group re-id, but also provides incremental improvement for single person re-id. As a result, the proposed MACG framework is capable of handling both group and person re-id tasks.

We visualize some person re-id results in Fig.~\ref{fig:vis}. It can be observed that for the single person re-id models (\ie, Strong-Baseline, PCB, and OSNet), it is difficult to distinguish people sharing similar appearances, especially in the cases of occlusions and illumination changes. Therefore, the Rank-1 matching accuracy is relatively low. In these cases, the proposed MACG model reduces the visual ambiguity by making use of group context information, and thus achieves better performance compared to single person re-id models.

\begin{table}[]
\centering
\caption{Dataset transfer results for single person re-id.}
\label{tab:transfer}
\renewcommand{\arraystretch}{1.5}
\begin{tabular}{l|cc|cc}
\hline
\multirow{2}{*}{Model} & \multicolumn{2}{c|}{CSG $\rightarrow$ Market-1501} & \multicolumn{2}{c}{CSG $\rightarrow$ MSMT17} \\ \cline{2-5} 
                       & mAP                     & R-1                    & mAP                  & R-1                  \\ \hline \hline
Strong-Baseline\cite{8930088}    & 15.9   &  35.0 &  2.9 & 8.1\\
PCB\cite{DBLP:conf/eccv/SunZYTW18} & 16.8  & 36.7 & 2.6  & 7.5                      \\
OSNet\cite{DBLP:conf/iccv/ZhouYCX19} & 18.8   &  38.4 &  2.7 & 7.8\\ \hline
MACG  & 14.9   &  39.2 &  2.0 & 8.3\\ \hline
\end{tabular}
\end{table}

To further evaluate the generalization capability of our framework for single person re-id, we directly apply the models trained on CSG to two single person re-id datasets, \ie, Market-1501~\cite{DBLP:conf/iccv/ZhengSTWWT15} and MSMT17~\cite{DBLP:conf/cvpr/WeiZ0018}. Since there does not exist group context on these two datasets, to build the context graph in our model, a single person is simply replicated to construct the group. The comparison results are illustrated in TABLE~\ref{tab:transfer}, where we can observe that the results of MACG are comparable to those models specifically designed for single person re-id~\cite{8930088,DBLP:conf/eccv/SunZYTW18,DBLP:conf/iccv/ZhouYCX19}. Notably, our model even achieves slightly higher Rank-1 accuracy on both datasets. This indicates that although our model is designed for group context learning, in the case of single person re-id where group context is missing, our model can still adaptively make use of the ``self-context'' from the replicated persons. As a result, the proposed MACG generalizes well to single person re-id without group context.

\begin{figure*}[t]
\subfloat[Backbone networks\label{subfig-2-1}]{%
   \includegraphics[width=0.25\linewidth]{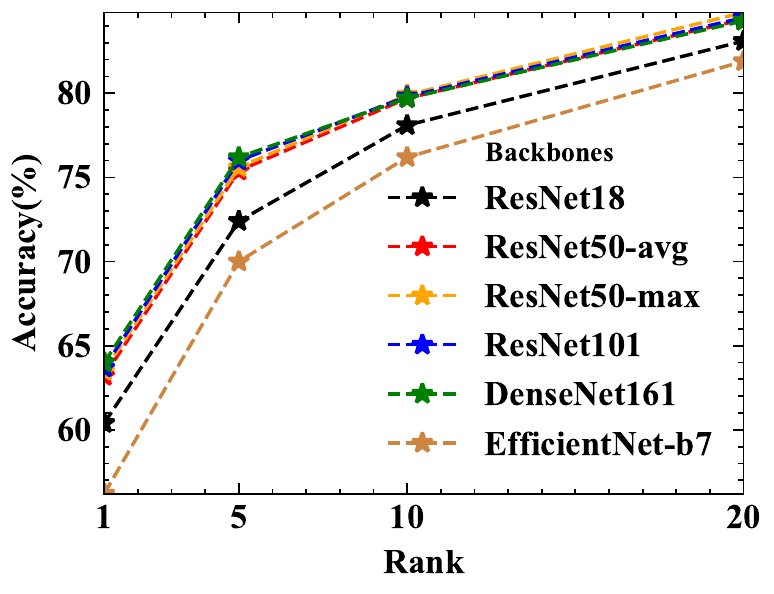}
}
\subfloat[Body height-parts\label{subfig-2-2}]{%
   \includegraphics[width=0.25\linewidth]{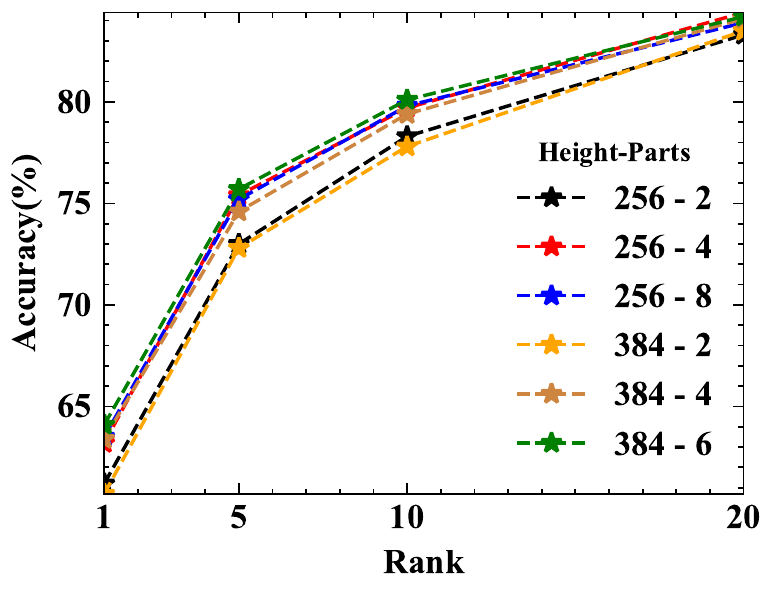}
}
\hfill
\subfloat[Node connections\label{subfig-2-3}]{%
   \includegraphics[width=0.25\linewidth]{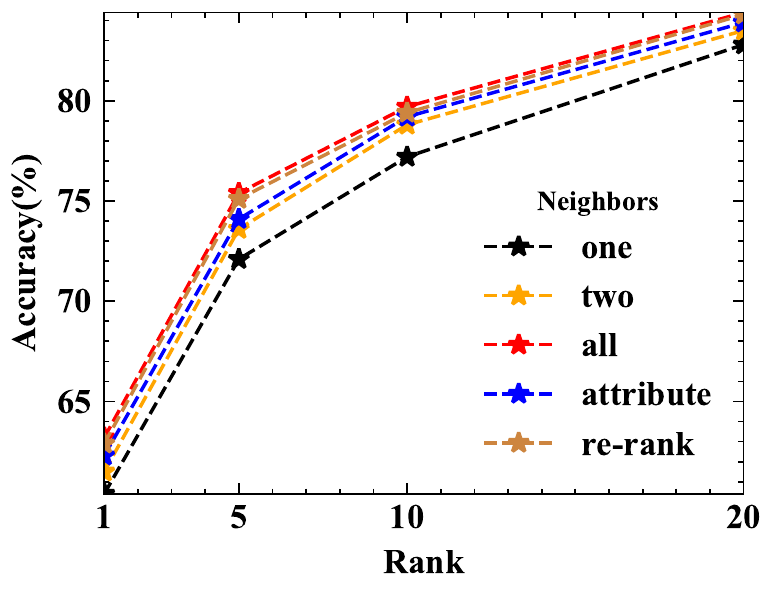}
}
\subfloat[GNN layers\label{subfig-2-4}]{%
   \includegraphics[width=0.25\linewidth]{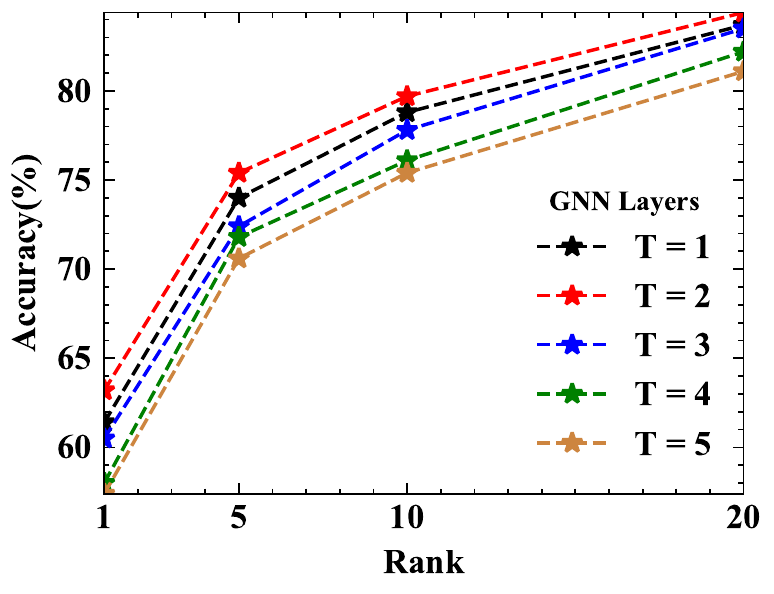}
}
 \caption{Model sensitivity analysis under different settings on the CSG dataset. Zoom in for better visualization.}
 \label{fig:sensitive}
\end{figure*}

\subsection{Model Component Analysis}
\begin{figure}[t]
    \centering
    \includegraphics[width=\linewidth]{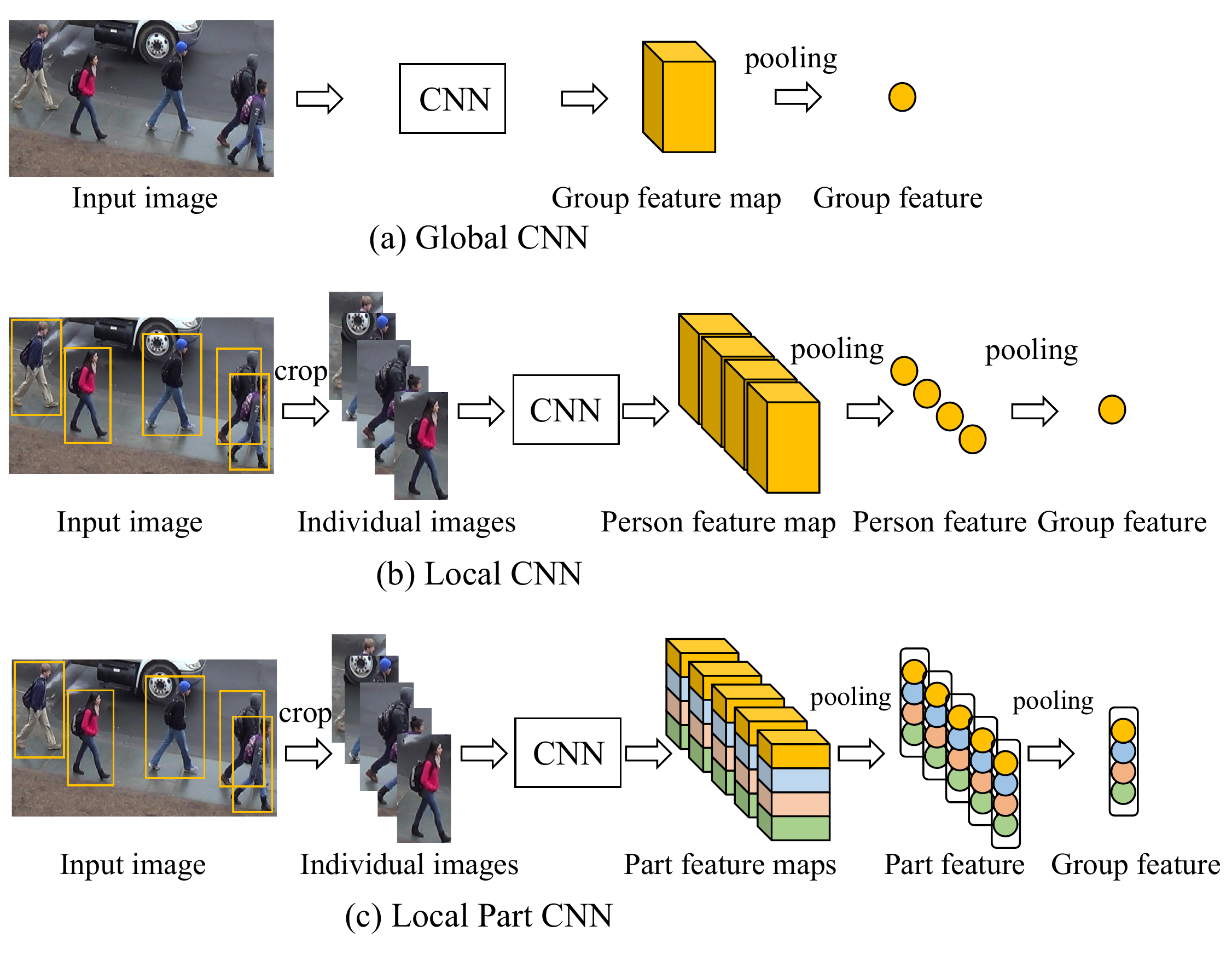}
    \caption{Illustration of three baseline CNN models.}
    \label{fig:baselines}
    \vspace{-2mm}
\end{figure}

The proposed MACG framework is built upon a GNN model with several attention modules, \ie, an intra-graph attention module, an inter-graph attention module, and a readout self-attention module. To validate the effectiveness of the GNN framework, we first compare our model with several baseline CNN models, which are illustrated in Fig.~\ref{fig:baselines}. The results are reported at the top of TABLE~\ref{tab:cmp2}. 1) We first feed the entire group image into the CNN model to directly learn the global group representation, which is denoted as `Global CNN' in the 1-st row of TABLE~\ref{tab:cmp2}. In this case, we observe very low performance on the CSG dataset. This is due to the fact that group images contain large portions of noisy background, which make it difficult for a single CNN to capture the group ID information. 2) We then crop the individual group members and train a single person re-id model to extract individual features. Mean-pooling is performed to yield the final group representation. This is denoted as `Local CNN' in the 2-nd row of TABLE~\ref{tab:cmp2}. The results are better than Global CNN since the influence of background is reduced, but they are still low due to the information loss during feature pooling. 3) As part models are effective at modeling individual people, we further train a part-based CNN model and concatenate the part features for the person-level representation. This is denoted as `Local Part CNN' in the 3-rd row of TABLE~\ref{tab:cmp2}, and the results are relatively better than the `Local CNN'. Note that both global and local CNN models cannot achieve satisfactory results on the group re-id task, because rich context information in groups has not been explored.

\begin{table}[t]
\centering
\caption{Model component analysis on the CSG dataset, where R-$k$ denotes the Rank-$k$ accuracy (\%).}
\label{tab:cmp2}
\renewcommand{\arraystretch}{1.5}
\begin{tabular}{l|c|cccc} \hline
   Model & Attention  & R-1 & R-5 & R-10 & R-20\\ \hline \hline
  Global CNN & - & 46.5& 64.3& 71.5& 77.1\\ 
  Local CNN &- & 52.7& 68.9& 74.5& 80.3\\ 
  Local Part CNN &- & 54.6& 70.1& 75.6& 80.9\\ \hline
  GNN &- & 58.7& 72.4& 76.3& 81.7\\
  Part GNN&- & 60.1& 73.2& 76.9& 82.5\\
  Part GNN& Intra-part& 61.3& 74.1& 77.5& 83.5\\
  Part GNN&Inter-part & 61.0& 74.3& 77.8& 83.3\\ 
  Part GNN&Intra-graph& 62.2& 74.5& 78.4& 83.8\\ 
  Part GNN& Inter-graph& 62.8& 74.8& 78.7& 84.0\\ 
  Part GNN & Readout& 61.2 & 73.9& 78.1& 82.9\\ \hline
  \textbf{MACG (Proposed)} & \textbf{Multi-level}  & \textbf{63.2}   & \textbf{75.4}  & \textbf{79.7} & \textbf{84.4} \\ \hline
\end{tabular}
\end{table}

\begin{figure*}[t]
    \centering
    \includegraphics[width=\linewidth]{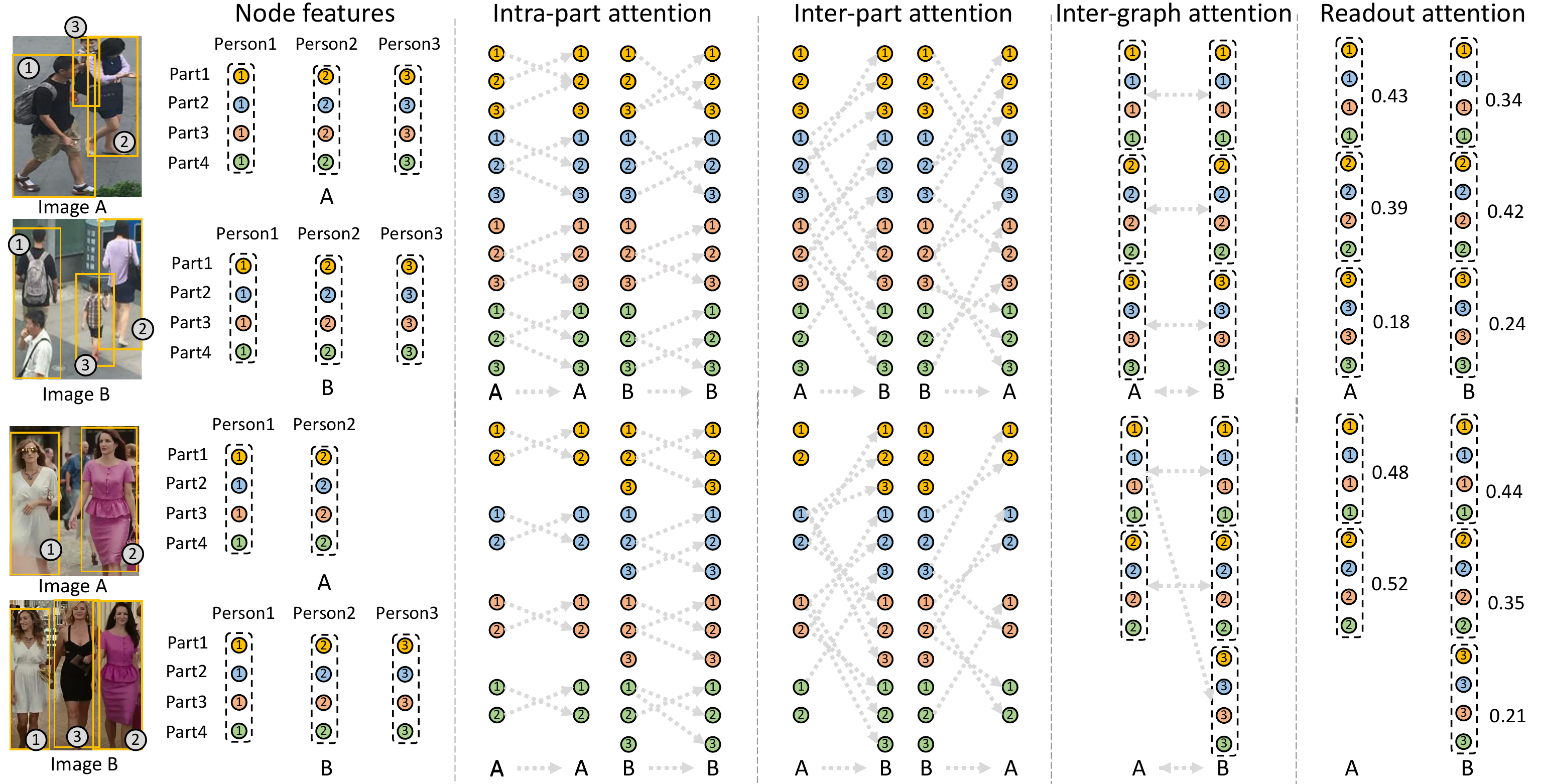}
    \caption{Visualization of the message passing path with multi-level attention mechanisms. Different colors denote different part features, and different numbers denote different people.}
    \label{fig:att}
\end{figure*}

To model context information, we first consider utilizing a vanilla GNN model for group representation learning, in which node features are extracted from the Local CNN.
From the 4-th row of TABLE~\ref{tab:cmp2}, we observe that the GNN model significantly improves the performance by 4\%, which demonstrates that modeling a group with a graph neural network can effectively capture the dependency within a group. We then replace the node features of the GNN with part-level features extracted by the local part CNN, reported as `Part GNN' in the 5-th row of TABLE~\ref{tab:cmp2}. The results are better than the GNN with person features, which demonstrates the importance of the node-level features for graph representation. For part representation, we further validate the effectiveness of the proposed intra-part and inter-part attention, as well as the intra-graph attention, by combining these two attention modules. It can be observed from the 6-th to 8-th rows of TABLE~\ref{tab:cmp2} that part-level attention can further improve the performance by exploring intra-group feature dependency. Moreover, inter-graph attention is proven to be effective from the 9-th row of TABLE~\ref{tab:cmp2}, as it explicitly models the correlation between group pairs. In addition, the readout attention is also effective for aggregating node features to yield the group representation, and we observe a 1.1\% improvement compared to the use of mean-pooling in Part GNN. Finally, by combining all the modules, our full model achieves 63.2\% rank-1 accuracy on the CSG dataset. The above results demonstrate the effectiveness of the proposed part-based graph model, as well as the attention modules.

\subsection{Model Sensitivity Analysis}
In this section, we analyze the sensitivity of the proposed framework under different settings and hyperparameters. 

1) We first analyze how different backbone networks influence model performance. Specifically, we replace the ResNet50 backbone with a shallower network (ResNet18) and a deeper network (ResNet101).  The results are illustrated in Fig.~\ref{subfig-2-1}. We observe considerable performance decrease when using shallow networks, especially for the rank-1 matching rate. Furthermore, ResNet101 only achieves negligible improvement compared with ResNet50. In addition to the ResNet family, we also evaluate the performance with DenseNet161~\cite{DBLP:conf/cvpr/HuangLMW17} and EfficientNet-b7
~\cite{DBLP:conf/icml/TanL19}. It can be observed that DenseNet161 achieves comparable performance with ResNet101, while EfficientNet-b7 performs worst among all these backbones, indicating that EfficientNet-b7 cannot be well generalized to our group-based re-id task. These results also indicate that different backbone networks have direct influences on the proposed framework. We find that ResNet50 is suitable for the backbone as it has a good trade-off between performance and model complexity.
We also compare different part feature pooling methods, \ie, average pooling and max pooling, based on the same backbone (\ie, ResNet50). From the performance of ResNet50-avg and ResNet50-max in Fig.~\ref{subfig-2-1}, we can see that pooling methods have limited influences on the final performance.

2) We then evaluate the influences of different image resolutions (\ie, human body heights) and human body partitions. To be specific, we resize the input person into 256/384$\times$128, and divide the human body into $P$ ($P$=2, 4, 6, 8) parts. The performance of different height-partition combinations is visualized in Fig.~\ref{subfig-2-2}. We can see that the performance of $P$=2 is notably inferior to other partitions, because this is a rather coarse partition. The best performance is achieved under the 384$\times$128 resolution and 6 partitions. It is also noteworthy that the performance of finer partitions (\eg, $P$=4, 6, 8) is close to each other, indicating that the model is not sensitive to human body partitions. 

3) In our framework, we construct the context graph by connecting all the members in a group. We compare it with partial connections according to the geometric distance between people. The results are reported in Fig.~\ref{subfig-2-3}, where `Neighbor=$K$' denotes that a person is only connected with his/her top-$K$ nearest neighbors and `Neighbor=All' denotes full connections. We observe that the performance is improved consistently as $K$ becomes larger, \ie, with denser connections. This indicates that a fully connected graph can facilitate message passing within the group, which helps the model to better learn the context information. We have also carried out more studies with respect to graph building and neighbor selection. In terms of graph building, we apply an offline attribute detector~\cite{DBLP:journals/corr/abs-2005-11909} on our proposed dataset, and we build the graph adjacency matrix according to the attribute similarities between persons. For neighbor selection, our current policy is to directly measure the correlations between nodes (Eq.~(\ref{eq:3}) and Eq.~(\ref{eq:9})), and we update the scores with the re-ranked similarity scores to evaluate the re-ranking policy~\cite{DBLP:conf/cvpr/ZhongZCL17}. We observe that neither attributes nor re-ranking policy improves the overall performance. On the one hand, as the attribute detector is trained on another source dataset, the domain gap will cause inaccurate predictions on the target dataset. On the other hand, since the re-ranking is only performed on limited samples between two groups, its impact is somewhat limited. 

4) Last but not least, we evaluate the influence of employing different numbers of GNN layers ($T = 1,2,3,4,5$). As shown in Fig.~\ref{subfig-2-4}, utilizing a two-layer GNN achieves better performance than employing a GNN with other numbers of layers. This is maybe because a single-layer GNN could not maintain sufficient representation capabilities, while a GNN with more layers contains too many parameters, making it more likely overfit to the training data. This may be related to the size of our training data, and further expanding the training data may require a deeper GNN to achieve better performance.



\subsection{Visualization of Attention Mechanisms}
To better understand how attention mechanisms work in our framework, we visualize the message passing path of each attention module, as illustrated in Fig.~\ref{fig:att}. In particular, each point in Fig.~\ref{fig:att} denotes a part-level feature, where different colors denote different parts and different numbers denote different people in the group. The arrow links the target node with its most attended neighbor. For intra-part attention, the model only considers features from the same body parts and thus there only exist connections between nodes of the same color. We observe that the occluded person is less involved in message passing. In the first example, the 3-rd person is partially occluded in \emph{Image A} and only two parts of this person are attended for intra-part message passing. 
Similarly, the bottom part of the first person in \emph{Image B} is occluded and thus is not attended.

\begin{table}[t]
\centering
\caption{Impact of detectors on the CSG dataset, where R-$k$ denotes the Rank-$k$ accuracy (\%).}
\label{tab:cmp-detector}
\renewcommand{\arraystretch}{1.5}
\begin{tabular}{l|l|c|cc} \hline
Group & Person & Recall & mAP & R-1 \\ \hline \hline
\multirow{6}{*}{GT} & GT &- &\textbf{62.3} & \textbf{63.2}  \\ 
 &  DPM\cite{DBLP:journals/pami/FelzenszwalbGMR10} &76.8 & 42.3 & 41.5 \\ 
 &  SSD\cite{DBLP:conf/eccv/LiuAESRFB16} &85.7& 54.6 & 54.2\\ 
 &  FasterRCNN\cite{DBLP:conf/nips/RenHGS15}  &94.8& 59.1& 59.9 \\ 
&  MaskRCNN\cite{DBLP:journals/pami/HeGDG20} &95.2& 59.5& 60.1 \\ 
 \hline
\multicolumn{2}{c|}{FasterRCNN\cite{DBLP:conf/nips/RenHGS15}}  &94.5 &  55.8 & 54.6 \\
\multicolumn{2}{c|}{Joint FasterRCNN\cite{DBLP:conf/cvpr/XiaoLWLW17} } &91.7 &  51.3 & 50.8\\
  \hline

\end{tabular}
\end{table}

For inter-part attention, the model tries to find dependencies between different body parts and thus only nodes in different colors are connected. We notice that the attended nodes come from adjacent parts of the target node. For example, the attended features of the 1-st part comes from the 2-nd part, and the attended features of the 2-nd part are from the 1-st and 3-rd parts, etc. This may be due to the fact that adjacent parts have stronger semantic connections with each other. For inter-graph attention, person-level attention weights are calculated, and we notice that the corresponding people successfully attend to each other. 
For readout attention, the attention weights are marked beside the features. We can see that the occluded persons are assigned with lower weights.
For the second example, there exists an outlier member in \emph{Image B}, and this sample is assigned with a relatively low attention weight. This demonstrates that readout attention is able to filter noisy sample in groups and thus yields more robust matching results.


\subsection{Discussion of Detectors}
\begin{figure}[t]
    \centering
    \subfloat[Pedestrian detection\label{fig:det_p}]{%
   \includegraphics[width=\linewidth]{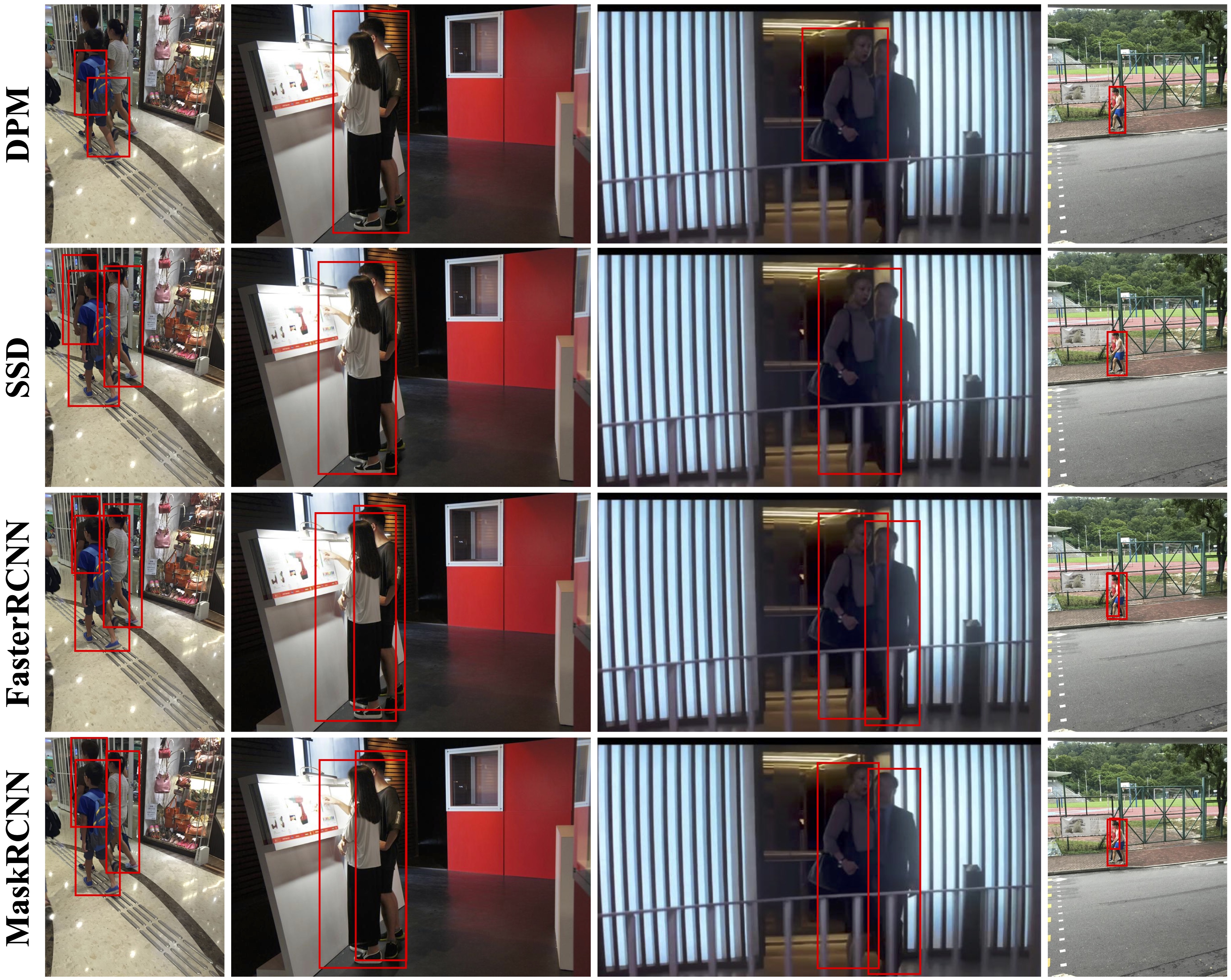}
}
    \vspace{1mm}
    \subfloat[Group and pedestrian detection\label{fig:det_g}]{%
   \includegraphics[width=\linewidth]{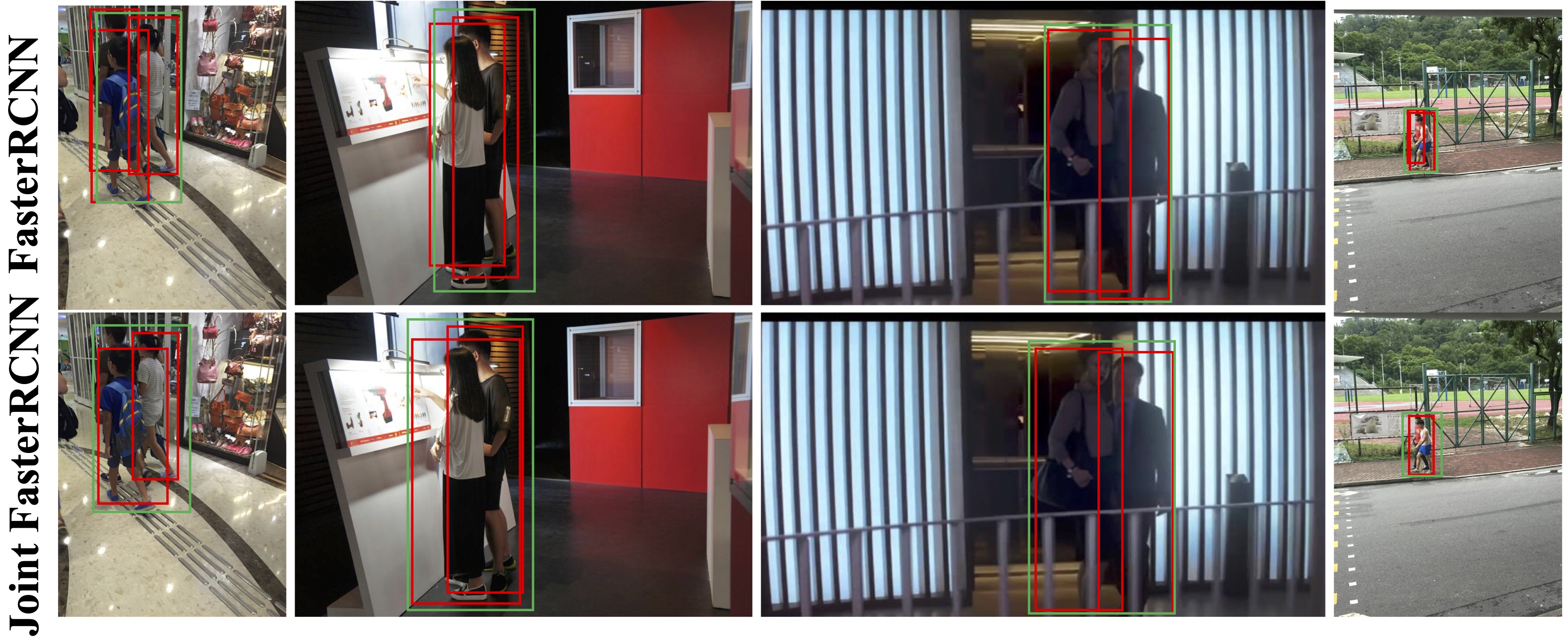}
}
    \caption{Visualization of detection results in the case of occlusion, low illumination and resolution. Red bounding boxes denote the detected pedestrians, and green bounding boxes denote the detected groups. Zoom in for better visualization.}
    \label{fig:det}
\end{figure}

In the above experiments, we utilize the ground-truth group and person bounding boxes for feature learning. To evaluate the impacts of pedestrian detectors, we employ different methods (\ie, DPM~\cite{DBLP:journals/pami/FelzenszwalbGMR10}, SSD~\cite{DBLP:conf/eccv/LiuAESRFB16}, FasterRCNN~\cite{DBLP:conf/nips/RenHGS15}, and MaskRCNN~\cite{DBLP:journals/pami/HeGDG20}) to generate pedestrian bounding boxes. It can be observed from TABLE~\ref{tab:cmp-detector} that the choice of the detector has a significant influence on recognition performance. We also show some detection examples in Fig.~\ref{fig:det_p}, where we set the positive detection threshold to -0.5 for DPM, and 0.5 for the other detectors. We can see that DPM is more sensitive to occlusions and scale/illumination variation as it is based on hand-crafted features, while deep learning based models (\ie, SSD, FasterRCNN, and MaskRCNN) are more robust in these situations and thus achieve better performance.

To explore the possibility of building an automatic group detector, we train a FasterRCNN~\cite{DBLP:conf/nips/RenHGS15} model which simultaneously detects individual pedestrians as well as groups based on our annotations. The detection results are illustrated in Fig~\ref{fig:det_g}, where candidate groups and pedestrians can be successfully localized. To further enable end-to-end optimization, we integrate the Joint FasterRCNN~\cite{DBLP:conf/cvpr/XiaoLWLW17} into our framework, \ie, the detection and recognition modules share the same backbone network, and the person-level features come from the RoI Pooling layer in FasterRCNN. We observe that this joint learning model achieves inferior performance than the two-stage FasterRCNN model. This is in consistence with recent two-stage person search frameworks~\cite{DBLP:conf/eccv/ChenZOYT18,Wang_2020_CVPR}, which indicate that different targets of detection and re-identification will influence the optimization of the joint learning framework. However, these results validate the possibility of building an end-to-end learning framework for group re-id.

\section{Conclusion and Future Work}\label{sec:conclusion}
In this work, we proposed a multi-attention context graph (MACG) model to jointly address the tasks of group re-id and group-aware person re-id. All the proposed attention mechanisms worked collaboratively for robust group representation. In addition, we built the CUHK-SYSU-Group (CSG) dataset for group re-identification, which is an order of magnitude larger than existing group re-id datasets. The superior performance in terms of the above two tasks on CSG and other group re-id benchmarks validated the effectiveness of the proposed MACG framework.

Currently, existing group re-id datasets do not contain any spatio-temporal information (\eg, time stamps or camera information). It would be interesting to combine such information for group re-id. In addition, the situation where there exists more than one group in an image is also worth future consideration, as the co-existing groups may provide additional context that is beneficial for group re-id.

\section*{Acknowledgment}
This work was supported partially by National Key Research and Development Program of China (2016YFB1001003), NSFC (U19B2035, 61527804, U1811461, 61976137, U1611461), STCSM (18DZ1112300), MoE-China Mobile Research Fund Project (MCM20180702), the 111 Project (B07022 and Sheitc No.150633), the Shanghai Key Laboratory of Digital Media Processing and Transmissions, and the SJTU-BIGO Joint Research Fund.
Yichao Yan and Jie Qin contributed equally to this work.




\ifCLASSOPTIONcaptionsoff
  \newpage
\fi



\bibliographystyle{ieee}
\bibliography{egbib}
%

%
\begin{IEEEbiography}[{\includegraphics[width=1in,height=1.25in,clip,keepaspectratio]{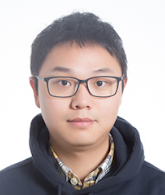}}]{Yichao Yan}
	received his B.E. and Ph.D degree in electrical engineering from Shanghai Jiao Tong University, in 2013 and 2019, respectively. Dr. Yan is currently a Research Scientist with the Inception Institute of Artificial Intelligence, UAE. He has authored/coauthored more than 10 peer-reviewed papers, including those in highly regarded journals and conferences such as TPAMI, CVPR, ECCV, ACMMM, IJCAI, TMM, etc. His research interests include object recognition, video analysis, and deep learning.
\end{IEEEbiography}

\begin{IEEEbiography}[{\includegraphics[width=1in,height=1.25in,clip,keepaspectratio]{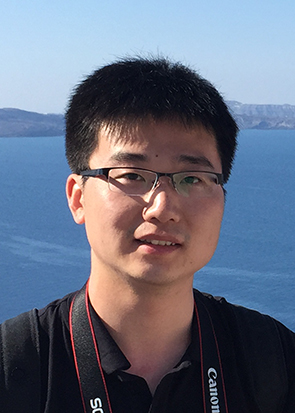}}]{Jie Qin}
	received the B.E. and Ph.D. degrees from Beihang University, China, in 2011 and 2017, respectively. From 2014 to 2015, he was a Visiting Researcher with The University of Sheffield, UK. From 2017 to 2018, he was a Postdoctoral Researcher with the Computer Vision Laboratory, ETH Z\"urich, Switzerland. He is currently a Research Scientist with the Inception Institute of Artificial Intelligence, UAE. His research interests include computer vision and machine learning. He has published more than 50 papers in main journals (TIP, TNNLS, TCSVT, PR) and top conferences (CVPR, ECCV, IJCAI, ACMMM). He has served as the Guest Editor for IJCV Special Issue on Efficient Visual Recognition and the Senior Program Committee Member for IJCAI 2020 and 2021.
\end{IEEEbiography}

\begin{IEEEbiography}[{\includegraphics[width=1in,height=1.25in,clip,keepaspectratio]{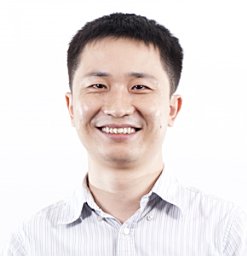}}]{Bingbing Ni}
received his Ph.D. from National University of Singapore (NUS), Singapore in 2011. Dr. Ni is currently a Professor in Shanghai Jiao Tong University.
His research interests are in the areas of computer vision, machine learning and multimedia.
He received the Best Paper Award from PCM'11 and the Best Student Paper Award from PREMIA'08.
\end{IEEEbiography}

\begin{IEEEbiography}[{\includegraphics[width=1in,height=1.25in,clip,keepaspectratio]{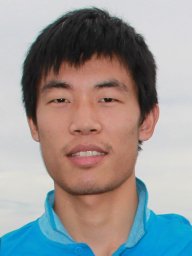}}]{Jiaxin Chen}
received the Ph.D degree in computer science and engineering from Beihang University, Beijing, China, in 2017. From 2017 to 2018, he was a Postdoctoral Researcher with the Multimedia and Visual Computing Laboratory, New York University Abu Dhabi, Abu Dhabi, UAE. He is currently a Research Scientist with the Inception Institute of Artificial Intelligence, UAE. His research interests include person re-identification, gait recognition, hashing and 3D shape analysis.
\end{IEEEbiography}

\begin{IEEEbiography}[{\includegraphics[width=1in,height=1.25in,clip,keepaspectratio]{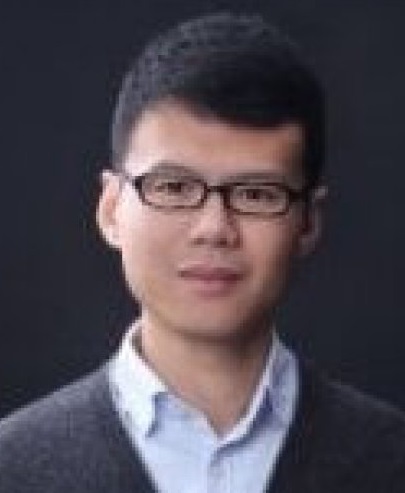}}]{Li Liu}
received the B.Eng. degree in electronic information engineering from Xian Jiaotong University, Xian, China, in 2011, and the Ph.D. degree from the Department of Electronic and Electrical Engineering, University of Sheffield, Sheffield, U.K., in 2014. He is currently the Director of Research in Computer Vision at Inception Institute of Artificial Intelligence, Abu Dhabi, UAE. His current research interests include computer vision, machine learning, and data mining.
\end{IEEEbiography}

\begin{IEEEbiography}[{\includegraphics[width=1in,height=1.25in,clip,keepaspectratio]{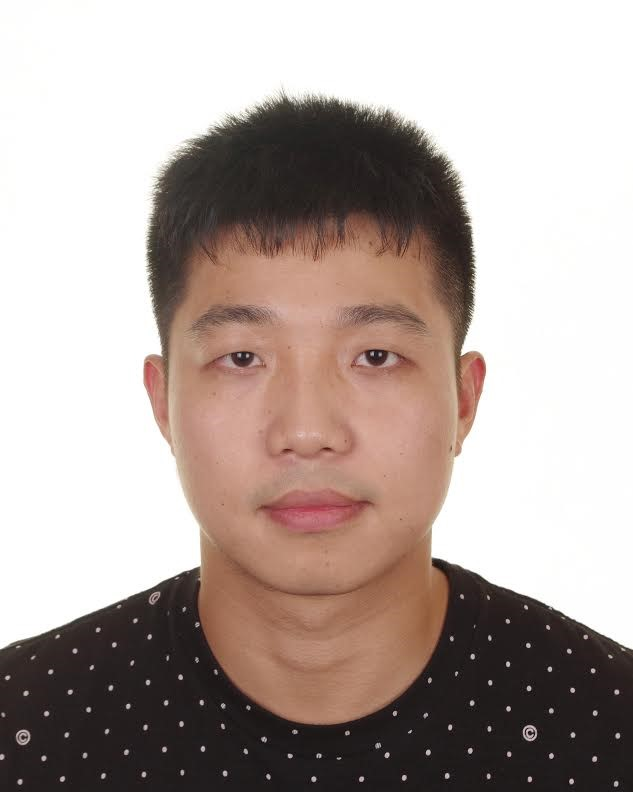}}]{Fan Zhu}
received his MSc degree in Electrical Engineering with distinction, Ph.D degree in computer vision from the University of Sheffield, UK, in 2011 and 2015 respectively. He was a post-doctoral research fellow with the Electrical and Computer Engineering department of New York University Abu Dhabi and a data scientist with Pegasus LLC. He is now the Director of Project Coordination at Inception Institute of Artificial Intelligence. His research interests include deep feature learning for 2D images and 3D shapes, scene understanding, video analytics and adversarial learning.
\end{IEEEbiography}

\begin{IEEEbiography}[{\includegraphics[width=1in,height=1.25in,clip,keepaspectratio]{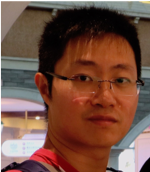}}]{Wei-Shi Zheng} is now a professor at Sun Yat-sen University. His research interests include person association and activity understanding in visual surveillance. He has now published more than 100 papers, including more than 70 publications in main journals (TPAMI, TNN, TIP, PR) and top conferences (ICCV, CVPR, IJCAI, AAAI). He served as an area chair for AVSS 2012, ICPR 2018 and BMVC 2018. He has joined Microsoft Research Asia Young Faculty Visiting Programme. He is a recipient of excellent youngscientists fund of the national natural science foundation of China, and a recipient of Royal Society-Newton Advanced Fellowship. He is an associate editor of the Pattern Recognition Journal.
\end{IEEEbiography}

\begin{IEEEbiography}[{\includegraphics[width=1in,height=1.25in,clip,keepaspectratio]{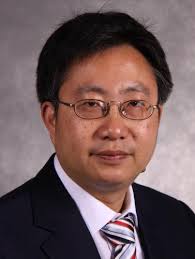}}]{Xiaokang Yang}
is currently a distinguished professor of School of Electronic Information and Electrical Engineering, and the deputy director of the Institute of Image Communication and Information Processing, Shanghai Jiao Tong University, Shanghai, China. He received the Ph.D. degree from Shanghai Jiao Tong University, Shanghai, China, in 2000.
He has published over 200 refereed papers, and has filed 60 patents. 
\end{IEEEbiography}

\begin{IEEEbiography}[{\includegraphics[width=1in,height=1.25in,clip,keepaspectratio]{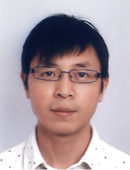}}]{Ling Shao}
is currently the CEO and the Chief Scientist of the Inception Institute of Artificial Intelligence, Abu Dhabi, United Arab Emirates. He is also the Executive Vice President and a Provost of the Mohamed bin Zayed University of Artificial Intelligence. His current research interests include computer vision, machine learning, and medical imaging. Dr. Shao is a fellow of IAPR, IET, and BCS. He is an Associate Editor of the IEEE TRANSACTIONS ON IMAGE PROCESSING, the IEEE TRANSACTIONS ON NEURAL NETWORKS AND LEARNING SYSTEMS, and several other top journals.
\end{IEEEbiography}






\end{document}